\documentclass[letterpaper,10pt,journal,twoside]{IEEEtran}

\usepackage{subfigure}
\usepackage{amsmath,amssymb,amsopn,amstext,amsfonts}
\usepackage[space]{cite}
\usepackage[table]{xcolor}
\usepackage{pifont}
\usepackage{multirow}
\usepackage{url}
\usepackage{caption}
\usepackage{adjustbox}
\usepackage{booktabs}
\usepackage{threeparttable}
\usepackage{makecell}
\usepackage{tabularx}
\usepackage{tikz}

\usepackage{array}

\usepackage{siunitx}
\pdfpxdimen=\dimexpr 1 in/72\relax

\newcommand{\etal}{\emph{et al}.}
\newcommand{\ie}{\emph{i}.\emph{e}.}
\newcommand{\eg}{\emph{e}.\emph{g}.}

\newcolumntype{L}[1]{>{\raggedright\arraybackslash}p{#1}}
\newcolumntype{Y}{>{\raggedright\arraybackslash}X}

\def\MYTITLE{Event-based SLAM Benchmark for High-Speed Maneuvers}

\makeatletter
\let\NAT@parse\undefined
\makeatother
\usepackage[pagebackref=false,breaklinks=true,colorlinks,citecolor=blue,linkcolor=blue,bookmarks=true,bookmarksnumbered=true]{hyperref}

\title{\MYTITLE}
\author{Sheng Zhong$^{\ast}$, Junkai Niu$^{\ast}$, Guillermo Gallego, Kaizhen Sun, Yang Yi, Zhiqiang Miao, \\ Dewen Hu, Yaonan Wang, Davide Scaramuzza, Yi Zhou\textsuperscript{\textdagger}  
\thanks{Sheng Zhong, Junkai Niu, Kaizhen Sun and Yi Zhou are with the Neuromorphic Automation and Intelligence Lab (NAIL) at the School of Artificial Intelligence and Robotics, Hunan University, Changsha, China. }
\thanks{Guillermo Gallego is with the Department of Electrical Engineering and Computer Science, Technische Universität Berlin, the Science of Intelligence Excellence Cluster, the Robotics Institute Germany and the Einstein Center Digital Future, Berlin, Germany.}
\thanks{Yang Yi and Dewen Hu are with the College of Intelligence Science and Technology, National University of Defense Technology, Changsha, China.}
\thanks{Zhiqiang Miao and Yaonan Wang are with the School of Artificial Intelligence and Robotics, Hunan University, Changsha, China.}
\thanks{Davide Scaramuzza is with the Robotics and Perception Group, Department of Informatics, University of Zurich, Switzerland.}
\thanks{$\ast$ equal contribution; \textdagger ~corresponding author (eeyzhou@hnu.edu.cn).}
}

\global\long\def\bfv{\mathbf{v}}

\definecolor{light-gray}{gray}{0.5}
\newcommand\gframe[1]{{\color{light-gray}\frame{#1}}}
\newcommand{\novalue}{{\textendash}}

\definecolor{lightgray9}{gray}{0.95}

\newcommand{\imgbox}[3]{
\begin{tikzpicture}
\node[anchor=south west, inner sep=0] (img)
    {\includegraphics[width=#3\linewidth]{#1}};
\begin{scope}[x={(img.south east)}, y={(img.north west)}]
\draw[red, line width=.75pt] #2;
\end{scope}
\end{tikzpicture}}

\begin{document}
\setcounter{figure}{-2}
\makeatletter
\g@addto@macro\@maketitle{
\vspace{1ex}
\def\figWidth{0.186\linewidth} 
    \centering
    \vspace{-0.2em}
    \setlength\tabcolsep{0pt}
    {\small
     \centering
                {\includegraphics[width=\linewidth]{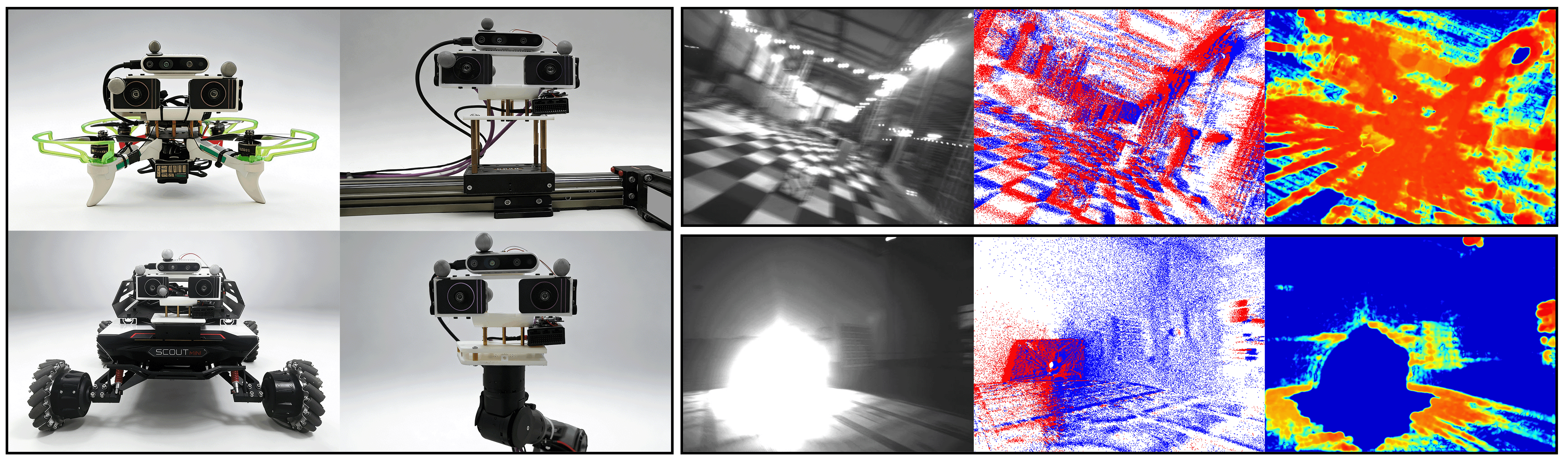}}
    }
\captionof{figure}{We present an event-based SLAM benchmark for high-speed maneuvering scenarios. 
Left: The multi-camera sensor suite is mounted on four mobile robotic platforms (drone, linear slider, robotic arm, and Mecanum-wheel vehicle) for data collection. 
Right: Visualizations of grayscale images, event streams (in red-blue according to event polarity), and optical flow magnitudes (pseudocolored as a heatmap) in high-speed maneuvering and HDR scenarios.}
\label{fig: eyecatcher}

\vspace{-3ex}
}
\makeatother
\maketitle


\begin{abstract}
Event-based cameras are bio-inspired sensors with pixels that independently and asynchronously respond to brightness changes at microsecond resolution, offering the potential to handle visual tasks in high-speed maneuvering scenarios.
Existing event-based approaches, although successful in mitigating motion blur caused by high-speed maneuvers, suffer from many limitations. 
Some of them highlight a success of pose tracking for a fronto-parallel fast shaking camera closed to the structure, while others assume pure (optionally aggressive) three-degree-of-freedom rotations. 
The former requires persistent local map visibility within the field of view (FOV), whereas the latter fails to generalize to six-degree-of-freedom (6-DoF) motions where both linear and angular velocities may be large.
Consequently, current successes do not fully demonstrate that event-based state estimation under arbitrary aggressive maneuvers is a fully solved problem.
To quantitatively assess the extent to which the potential of event cameras has been unlocked, we conduct a thorough analysis of state-of-the-art (SOTA) event-based visual odometry (VO)/visual-inertial odometry (VIO) methods and report shortcomings in current public datasets. 
Furthermore, we introduce a benchmarking framework for event-based state estimation, called EvSLAM, characterized by sufficient variation in data collection platforms, diverse extreme lighting scenarios, and a wide scope of challenging motion patterns under a clear and rigorous definition of high-speed maneuvers for mobile robots, along with a novel evaluation metric designed to fairly assess the operational limits of event-based solutions.
This framework benchmarks state-of-the-art methods, yielding insights into optimal architectures and persistent challenges.
\end{abstract}
\section*{Multimedia Material}
\noindent Video: {\small \url{https://youtu.be/VW6gsWlqOaM}}\\
Dataset: {\small\url{https://github.com/NAIL-HNU/EvSLAM_Dataset}}

\section{Introduction}
\label{introduciton}
Visual Simultaneous Localization and Mapping (vSLAM) has served as a critical technology for a wide range of robotic applications~\cite{Cadena16tro, Macario22survey}, such as autonomous driving, extended reality (XR) devices, and robotic navigation. 
Despite their reliable performance in most scenarios, mainstream frame-based vSLAM algorithms are challenged by conditions like high dynamic range (HDR) and agile/aggressive motion, where under/over-exposure and motion blur degrade image quality and, consequently, impair algorithm performance.
The emergence of event cameras offers a promising alternative to address these limitations.

Event cameras \cite{Gallego20pami,Posch14ieee}
are bio-inspired sensors that report logarithmic brightness changes independently per pixel, called ``events''. 
This operating principle avoids capturing full images at a fixed rate, thereby reducing both power consumption and bandwidth requirements. 
More importantly, event cameras possess microsecond-level temporal resolution and a dynamic range larger than 120 dB~\cite{Lichtsteiner08ssc}, making them well-suited for the above-mentioned scenarios that pose significant challenges to frame-based cameras.

Although event-based SLAM algorithms show great promise~\cite{Gallego25slam,Rebecq18ijcv,Rebecq17ral,Rosinol18ral,Guo24tro}, they are not yet mature enough to fully realize their potential.
One key reason is that the unique asynchronous output of event cameras is incompatible with existing modules in VO/SLAM pipelines designed for standard cameras, necessitating the exploration of new paradigms compatible with the event-based modality. 
Another major challenge is data scarcity, i.e., the shortage of public, real-world event datasets.

Over the past decade, the development of event cameras has led to the emergence of several datasets, which in turn have effectively fostered the advancement of event-based SLAM algorithms.
However, most datasets provide only a small number of challenging HDR sequences and suffer from shortcomings.
Datasets that claim to offer high-speed motion scenarios are often confined to back-and-forth motion in small-scale environments~\cite{Gao22ral,Klenk21iros} or exhibit limited relative motion due to excessively large scene depth~\cite{Chaney23cvprw}. 

This paper proposes an event-based SLAM benchmark featuring high-speed maneuvering scenarios. 
This benchmark comprises multiple sequences recorded on various mobile robotic platforms using a stereo event camera (DVXplorer: 640$\times$480 px resolution) with a baseline of 10 cm, covering a wide scope of challenging motion patterns and diverse extreme lighting scenarios.
We employ a simple and accurate lens-focusing strategy to address the spatio-temporal stereo inconsistency of event data found in some public datasets.
An integrated inertial measurement unit (IMU) provides 3-axis accelerometer and gyroscope measurements at 400 Hz.
The benchmark also includes synchronous stereo grayscale images and monocular color images at the same pixel resolution, thus enabling research on event-image fusion and evaluation with state‑of‑the‑art frame‑based methods. 
Moreover, we conduct a comprehensive review and analysis of the latest advances in event-based VO/SLAM methods. 

The contribution of the paper is summarized as follows:
\begin{itemize}
\item We provide a comprehensive review of recent developments in event-based VO/SLAM across various methodological categories, with a critical analysis of their characteristics, limitations and future improvements (Sec.~\ref{sec: related work}).
\item We report shortcomings of current datasets, including: spatio-temporal inconsistency across stereo observations, the lack of comprehensive high-speed maneuvering scenarios, and the computational burden caused by the increasing pixel resolution (Sec.~\ref{sec: imperfections}).
\item We provide sufficient variation in data collection platforms, incorporating diverse extreme lighting scenarios, and covering a wide scope of challenging motion patterns under a clear and rigorous definition of high-speed maneuvers for mobile robots (Sec.~\ref{sec: dataset}).
\item We support our discussion through a comprehensive benchmarking of all state-of-the-art solutions using the proposed dataset and evaluation metrics (Sec.~\ref{sec: benchmark}). 
\end{itemize}

By combining a review of prior art with an evaluation of representative methods on the proposed benchmark, we facilitate a meaningful discussion of future research directions in the field.
We believe this integrated effort will contribute to advancing the real-world application of event cameras in high-speed maneuvering scenarios.

\section{Related Work}
\label{sec: related work}
The high temporal resolution and high dynamic range of event cameras provide the potential to handle high-speed maneuvering and extreme lighting conditions. 
Over the past decade, with the development of event camera hardware and the mutual promotion of event-based SLAM algorithms and datasets, this potential has been gradually unlocked. 
To better assess the extent to which the potential of event cameras has been realized, we conduct a literature review on the latest advances in event-based SLAM methods and datasets.

\subsection{Event-based VO/SLAM Methods}
\label{subsec: methods}
\begin{figure}[t]
	\centering

    {\includegraphics[trim={3.2cm 0.3cm 8cm 7.3cm},clip,width=\linewidth]{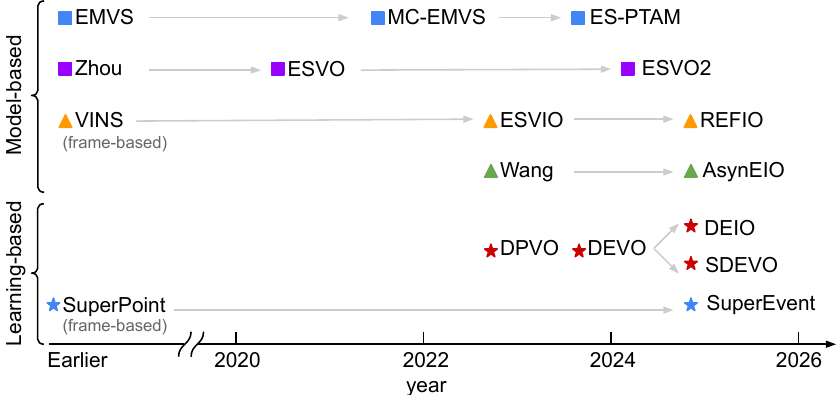}}

	\caption{\textit{Evolution of some event-based SLAM methods.} 
    Identical color indicates methods obtained by improving a common original design, 
    while identical shape denotes methods belonging to the same category: Direct (square), Indirect (triangle), or Learning-based (star).} 
    \label{fig: method tree}
\end{figure}

In recent years, event-based SLAM methods have witnessed considerable development. 
Given that a significant number of survey papers~\cite{Gallego20pami,Huang24arxiv,Tenzin24survey,Ghosh24survey,Gallego25slam} have already presented comprehensive reviews of this field, our review focuses on the recent advances in different categories of event-based SLAM algorithms, primarily analyzing their characteristics and fundamental limitations, coupled with an analysis of their performance in high-speed maneuvering scenarios, to explore the future direction of this field.
Table~\ref{tab:methods_comparison} provides a simple categorization of the surveyed methods and Fig.~\ref{fig: method tree} illustrates the evolution of some of these methods.

\subsubsection{Model-based Methods}
We begin by reviewing model-based methods, which estimate the camera trajectory by processing events and explicitly constructing photometric and geometric constraints. 
From the perspective of event data processing, existing model-based methods can be broadly categorized into direct and feature-based (i.e., indirect) methods.

\begin{table}[t]
    \centering
    \begin{adjustbox}{max width=\linewidth}
    \begin{threeparttable}
        \caption{\emph{Comparison of different event-based SLAM methods.}\label{tab:methods_comparison}}

\renewcommand{\arraystretch}{1.2}
\rowcolors{1}{}{lightgray9}
\begin{tabular}{llllll}
\toprule
\textbf{Method}  & \textbf{M/L} & \textbf{D/I} &  \textbf{Event Representation} & \textbf{BA} & \textbf{Input}  \\
\midrule
ESVO~\cite{Zhou20tro} & M  & D & Time Surface (TS) &\ding{55} & 2E \\
EDS~\cite{Hidalgo22cvpr} & M  & D & Event Image &\ding{51} & E+F \\
Wang et al.~\cite{Wang23ral} & M  & I & Binary Image &\ding{51} & 2E \\
ESVIO~\cite{Chen23ral} & M  & I & Time Surface &\ding{51} & 2E+2F+I \\
ESVO\_AA~\cite{Niu24icra} & M & D & Time Surface &\ding{55} & 2E+I \\
PL-EVIO~\cite{Guan24tase} & M & I & Time Surface &\ding{51} & E+F+I \\
DEVO~\cite{Klenk24threedv} & L & I & Event Voxel Grid &\ding{51} & E \\
ES-PTAM~\cite{Ghosh24eccvw} & M & D & Events &\ding{55} & 2E \\
ESVO2~\cite{Niu25esvo2} & M & D & Time Surface &\ding{55} & 2E+I \\
DEIO~\cite{Guan25iccvw} & L & I & Event Voxel Grid &\ding{51} & E+I \\
EVIV~\cite{Lu25tro} & M & I & Events  &\ding{55} & 2E+I \\
SuperEvent~\cite{Burkhardt25iccv} & L & I & Multi-channel TS &\ding{51} & E \\
SDEVO~\cite{Zhong25ral} & L & I & Event Voxel Grid &\ding{51} & 2E \\
AsynEIO~\cite{Wang25asyneio} & M & I & Time Surface &\ding{51} & 2E+I \\
\bottomrule
\end{tabular}
    The columns specify, from left to right: the methodological category (\textbf{M}odel-based or \textbf{L}earning–based; 
    \textbf{D}irect or \textbf{I}ndirect); 
    the event representation; whether the method has a global refinement module; 
    and the sensor modalities used (\textbf{E}vent camera, \textbf{F}rame-based camera, \textbf{I}MU), 
    where ``2'' denotes two sensors (i.e., stereo).
    \end{threeparttable}
\end{adjustbox}
\end{table}
\paragraph{Direct Methods}
Direct methods refer to approaches that directly process raw sensor data (i.e., pixel intensities or events), typically through some intermediate representation.
Leveraging the edge-triggered and high-temporal-resolution properties of event cameras, Rebecq \etal{} propose EMVS~\cite{Rebecq18ijcv}, which projects monocular event streams as 3D rays and reconstructs semi-dense 3D structure by searching local maxima of ray density. 
Building on this, Ghosh \etal{} present \mbox{ES-PTAM}~\cite{Ghosh24eccvw}, which adopts a stereo event setup to fuse cross-view rays for higher accuracy, and integrates an edge-map alignment module to align 3D structures with binary event maps for joint camera pose optimization.

Zhou \etal~\cite{Zhou18eccv} present a different mapping method for reconstructing a semi-dense 3D map of the scene, which is designed based on the assumption of spatio-temporal consistency of events across stereo image planes, as measured using time surfaces (TS) (Tab.~\ref{tab:methods_comparison}).
This mapping module is combined with a camera tracking module, based on 2D-3D registration between the TS and the map to form a VO system called ESVO~\cite{Zhou20tro}. 
Niu \etal~extend this work by incorporating IMU data for tightly-coupled optimization and utilizing an efficient edge pixel sampling strategy to propose ESVO2~\cite{Niu25esvo2}, achieving higher accuracy and efficiency across multiple datasets.


Benefiting from accurate 3D reconstruction, all the aforementioned methods achieve precise and even high-frequency tracking results in scenarios with clear environmental textures and stable camera motion. 
However, the true potential of event cameras lies in handling challenging conditions, such as HDR and high-speed maneuvering scenarios, for which current direct methods are fundamentally limited. 
The significantly increased ratio of noisy events and more discrete distribution of signal events in such extreme environments degrade both the accuracy and the quantity of reconstructed edge textures, consequently leading to a decline in tracking precision. 
Moreover, under high-speed maneuvers, the increased convergence distance required for 2D-3D registration significantly raises the probability of the optimization problem becoming trapped in zero-gradient regions, resulting in camera tracking failure, especially as event representations inherently lack the rich content found in intensity images.

\paragraph{Indirect Methods}
These methods involve detecting and tracking features (e.g., keypoints, lines) within the event stream, followed by motion estimation using the geometric constraints provided by such features.
To date, most of these methods~\cite{Alzugaray18ral,Li19iros,Hadviger21ecmr} are adaptations of techniques originally designed for frame-based cameras~\cite{Harris88,Rosten06eccv}. 
In 2023, Chen \etal{} proposed ESVIO~\cite{Chen23ral} by incorporating events into the VINS-Fusion~\cite{Qin18tro} framework. 
This method detects features using the ARC*~\cite{Alzugaray18ral} algorithm on motion-compensated event representations and tracks them via Lucas-Kanade optical flow~\cite{Lucas81ijcai}. 
By performing tightly-coupled optimization with IMU data, the system achieves robust pose estimation and supports both events-and-frames and events-only modes.

Despite the success of feature-based methods like ESVIO on certain datasets, the inherent motion-dependent nature of event data makes them fundamentally incompatible with feature tracking techniques designed for frame-based cameras.
Factors such as noise, abrupt changes in motion direction, and even speed variations can severely compromise the temporal consistency of features.
To address this limitation, some works exploit the asynchronous nature of event sensors to continuously track features in the event streams. 
For example, Alzugaray \etal{} proposed HASTE \cite{Alzugaray20bmvc}, a method that asynchronously tracks patch-wise features using only events, continuously updating feature states by selecting optimal motion hypotheses to maximize correlation between events and template patches. 
Extending this work, Ikura \etal{} proposed RATE \cite{Ikura24rate}, which integrates Shi-Tomasi~\cite{Shi94cvpr} corner detection on binary event frames for automatic feature initialization. 
More recently, Wang \etal~\cite{Wang25asyneio} utilize a front-end capable of event-by-event tracking to output continuous feature trajectories. 
By integrating these with inertial measurements within a Gaussian regression framework for optimization, which was used in~\cite{Wang23ral}, they achieve continuous-time state estimation and demonstrate considerable accuracy across multiple datasets.

Although these methods exploit the asynchronous nature of events, two critical problems persist. 
First, current continuous feature tracking strategies are heavily based on hand-crafted rules, raising doubts about their stability and generalization. 
Second, processing events individually while maintaining multiple trackers imposes a significant computational burden.
Given that these methods do not demonstrate the potential to surpass the accuracy of learning-based methods, it is questionable whether such an overhead is justifiable for SLAM.

\subsubsection{Learning-based Methods}
Unlike model-based approaches, learning-based methods utilize deep networks to detect implicit information from event streams or representations for feature tracking and pose estimation.
As early as 2018, Zhu \etal{} \cite{Zhu19cvpr} and Ye \etal{} \cite{Ye19arxiv} employed convolutional neural networks to predict optical flow, semi-dense depth and ego-motion via unsupervised learning. 
However, only recently have data-driven deep learning methods demonstrated significantly higher accuracy and sufficient generalization capabilities compared to model-based methods.

Adapting the network architecture of frame-based method DPVO~\cite{Teed23dpvo} to the event modality, Klenk \etal~\cite{Klenk24threedv} proposed DEVO. 
This method introduces a dedicated patch selection mechanism and is trained on simulated event data, achieving superior performance on a variety of real-world datasets. 
Subsequent methods~\cite{Guan25iccvw,Zhong25ral} incorporate additional sensors to resolve scale uncertainty in DEVO while further improving tracking accuracy.

Another common approach involves performing feature detection and description on event representations followed by pose estimation based on geometric constraints. 
Leveraging the pixel-alignment between events and frames provided by the DAVIS camera \cite{Brandli14ssc}, Yannick \etal~\cite{Burkhardt25iccv} generate sparse pseudo-labels for training using frame-based feature detection~\cite{Detone18cvprw} and matching~\cite{Sarlin20cvpr} techniques. 
By combining an information-rich event representation with a transformer backbone, SuperEvent~\cite{Burkhardt25iccv} predicts stable keypoints with expressive descriptors. 

\begin{table*}[t!]
\centering
\caption{\label{tab:dataset comparison}\emph{Comparison of several event camera datasets suitable for SLAM tasks.}
    The columns ``Pixels'' and ``Baseline'' refer specifically to the event camera. 
    Sensor notation is consistent with the previous table.}
\setlength{\tabcolsep}{4pt}
\begin{adjustbox}{max width=\linewidth}
\renewcommand{\arraystretch}{1.2}
\rowcolors{1}{}{lightgray9}
\begin{tabular}{llrllll}
\toprule
\textbf{Dataset}  & \textbf{Pixels} & \textbf{Baseline} & \textbf{Sensors} & \textbf{Platform} & \textbf{Terrain} & \textbf{Ground Truth Pose} \\
\midrule
ECDS~\cite{Mueggler17ijrr} & 240$\times$180  & - & E, I & Handheld & Indoor, Outdoor & MoCap (partial) \\
RPG~\cite{Zhou18eccv} & 346$\times$260 & 14.7 cm & 2E, I  & Handheld & Indoor & MoCap \\
UZH-FPV~\cite{Delmerico19icra} & 346$\times$260 & -  & E, 2F, I  & Drone & Indoor, Outdoor & Laser Tracker + MoCap \\
MVSEC~\cite{Zhu18ral} & 346$\times$260 & 10 cm & 2E, 2F, I, Lidar, GPS  & Handheld, Car, Drone, Motorcycle & Indoor, Outdoor & LiDAR-SLAM + MoCap \\
DSEC~\cite{Gehrig21ral} & 640$\times$480  & 60 cm & 2E, 2F, I, Lidar, GPS  & Car & Suburban & \novalue \\
TUM-VIE~\cite{Klenk21iros} & 1280$\times$720 & 11.84 cm & 2E, 2F, I & Helmet, Handheld & Indoor, Outdoor & MoCap (partial) \\
M2DGR~\cite{Yin22ral} & 640$\times$480 & - &  E, F, I, D, Lidar, GPS & Ground Robot & Indoor, Outdoor & MoCap + GNSS-RTK \\
EDS~\cite{Hidalgo22cvpr} & 640$\times$480 & - &  E, F, I & Handheld & Indoor & MoCap \\
VECtor~\cite{Gao22ral} & 640$\times$480 & 17 cm &  2E, 2F, I, D, Lidar & Helmet, Cart & Indoor & ICP + MoCap \\
FusionPortable~\cite{Jiao22iros} & 346$\times$260 & 25 cm &  2E, 2F, I, Lidar, GPS & Handheld, Quadruped, Vehicle & Indoor, Outdoor & Mocap + 6DoF NDT \\
HKU~\cite{Chen23ral} & 346$\times$260 & 6 cm & 2E, 2F, I & Drone & Indoor & MoCap \\
M3ED~\cite{Chaney23cvprw} & 1280$\times$720 & 12 cm & 2E, 2F, I, Lidar, GPS  & Car, Drone, Quadruped & Urban, Forest & LiDAR-SLAM + GNSS-RTK \\[0.8pt]
ECMD~\cite{Chen23ecmd} & 640$\times$480, 346$\times$260 & 30 cm  & 4E, 3F, I, 3Lidar, GPS  & Car & Urban & INS + GNSS-RTK \\
FusionPortableV2~\cite{Wei24ijrr} & 346$\times$260 & 25/73 cm &  2E, 2F, I, Lidar, GPS & Handheld,  Car, Quadruped, Vehicle & Indoor, Outdoor & Laser Tracker + GNSS-RTK \\
CoSEC~\cite{Peng24arxiv} & 1280$\times$720 & -  & 2E, 2F, I, Lidar, GPS  & Car & Suburban &  LiDAR-SLAM \\
Ours & 640$\times$480 & 10 cm & 2E, 2F, I & Slider, Drone, Robotic arm, Vehicle & Indoor & MoCap  \\
\bottomrule
\end{tabular}
\end{adjustbox}
\end{table*}
The aforementioned state-of-the-art data-driven methods have been tested on our proposed high-speed maneuvering dataset (Sec.~\ref{sec: benchmark}), either directly or by integrating them with classic VIO frameworks. 
These methods achieve considerably accurate results in pose estimation, demonstrating their capability to handle high-speed maneuvering scenarios common in mobile robotics.
Furthermore, these methods even exhibit superior generalization compared to model-based approaches, which often require extensive parameter tuning when adapting to unseen environments.
Therefore, at present, data-driven learning-based methods remain the most promising approaches for unlocking the potential of event cameras in VO/SLAM, particularly in addressing extremely challenging scenarios.

However, these methods still face several limitations and challenges that hinder their practical deployment and future development.
First, due to the complexity of event representations and the large scale of current network architectures, existing approaches typically achieve real-time performance only on high-end GPUs, making them unsuitable for embedded platforms and other edge computing devices. 
Addressing this limitation will require more efficient and lightweight network designs and hardware.
Second, most data-driven methods continue to follow the conventional pipeline developed for frame-based cameras, where the front-end produces feature trajectories and the back-end fits them using classical VIO optimization. 
Such a design fails to fully exploit the high temporal resolution and asynchronous nature of event data. 

In this regard, EVIV~\cite{Lu25tro}, a map-free design for event-based visual-inertial velocity estimation, provides a compelling alternative. 
Lu \etal{} employ stereo event streams and IMU to explicitly estimate normal flow (i.e., optical flow along the image gradient) via local differential methods, while deriving depth and uncertainty in the front-end. 
Their back-end fuses normal flow and IMU data in a continuous-time framework to estimate linear velocity and IMU biases, achieving accurate velocity estimation in extreme high-speed maneuvers (e.g., rapid vehicle spinning), where most methods fail. 
Unlike the computationally expensive continuous-time back-end in \cite{Wang25asyneio} (even on GPU), EVIV outputs body-frame linear velocity at 80 Hz for 346$\times$260 px event data.
Building on this insight, we suggest that while learning-based VIO should continue to explore the potential of event cameras for high-frequency feature tracking, it is also worthwhile to pursue more efficient and precise estimation of local event representations containing motion information, such as normal flow.
Correspondingly, an efficient continuous‑time optimization back-end capable of integrating asynchronous constraints with inertial measurements is also an essential component for future development.

\subsection{Event-based SLAM Datasets}
\label{sec: datasets}

Over the past decade, numerous datasets combining event cameras with other sensing modalities have flourished, significantly facilitating the development of algorithms in this field. 
This section presents a concise review of common datasets used for SLAM tasks, as shown in Tab.~\ref{tab:dataset comparison}.

In 2014, Weikersdorfer \etal~\cite{Weikersdorfer14icra} proposed one of the earliest cross-modal datasets that integrates event data (128$\times$128 px resolution) and RGB-D data, capturing indoor sequences with ground truth poses provided by a motion capture (MoCap) system.
Later, Mueggler \etal~\cite{Mueggler17ijrr} utilized a DAVIS240C camera to provide synchronized event streams, grayscale images and IMU data. 
The sequences primarily involved 6-DoF motions in front of various high-textured indoor scenes.

Driven by advances in event camera hardware and algorithms, higher-resolution event cameras and more sophisticated sensor configurations have emerged. 
The MVSEC~\cite{Zhu18ral} dataset is the first cross-modal dataset to feature a comprehensive sensor suite, which includes a pair of mDAVIS event cameras (346$\times$260 px resolution) with a 10 cm baseline, a VI-Sensor (stereo frame cameras and 9-axis IMU) and a 16-channel \mbox{LiDAR}. 
It captures sequences exhibiting diverse motion patterns across multiple platforms, including a car, a motorcycle, a handheld setup, and a hexacopter.
Ground truth for indoor sequences is provided by a MoCap system, while for outdoor sequences it is obtained by fusing \mbox{LiDAR} and IMU data using Cartographer~\cite{Hess16icra}.
The FusionPortable~\cite{Jiao22iros} dataset adopts a similar sensor configuration, while incorporating novel mobile platforms, such as a quadruped robot. 
Its follow-up, FusionPortableV2~\cite{Wei24ijrr}, further introduces car-based data collection and expands the diversity and scale of outdoor scenarios.

Since 2021, a growing number of datasets with similar configurations have been introduced~\cite{Yin22ral}.
Gehrig \etal present DSEC~\cite{Gehrig21ral}, the first high resolution large-scale dataset with stereo event cameras, which employs two Prophesee Gen~3 sensors with a baseline of 60 cm, increasing the event resolution to 640$\times$480 px.
The sensor suite is further equipped with stereo frame cameras, a GNSS receiver, and a \mbox{LiDAR} unit that is used to generate ground truth depth maps. 
This configuration captured numerous driving scenarios under various illumination conditions in suburban areas.
Following the DSEC dataset, both the ECMD~\cite{Chen23ecmd} and CoSEC~\cite{Wei24ijrr} datasets focus on driving scenarios; 
ECMD is distinguished by a rich sensor configuration, particularly featuring two pairs of event cameras with different resolutions and two infrared cameras, recording extensive urban sequences.
Gao \etal~\cite{Gao22ral} focus on indoor scenarios, employing multiple platforms including a helmet-mounted rig and a pushcart equipped with stereo VGA-resolution (Video Graphics Array) event cameras, stereo industrial cameras, and a \mbox{LiDAR}. 
This dataset, VECtor, contains sequences captured in rooms and corridors, with some involving relatively high-speed motion, and all of these sequences include loop closures. 
Their ground truth poses are provided by a combination of MoCap and \mbox{LiDAR}-SLAM.
Notably, driven by the focus on algorithms that fuse event data and image frames, Hidalgo-Carrio \etal~\cite{Hidalgo22cvpr} employ a beamsplitter to enable a DVXplorer event camera and a VGA-resolution color FLIR camera to observe the same scene simultaneously. 
Using this setup, they record a monocular event-based dataset that overcomes the limitation where high-resolution event cameras cannot directly capture image frames.

Furthermore, Klenk \etal~\cite{Klenk21iros} introduce TUM-VIE, the first dataset achieving an event resolution of 1~Megapixel (HD), with a sensor configuration comprising two Prophesee Gen~4 (1280$\times$720 px) event cameras and two industrial cameras.
They collect data both indoors and outdoors using handheld and helmet-mounted setups. 
Chaney \etal~\cite{Chaney23cvprw} introduce the M3ED dataset, recording extensive challenging scenes—such as urban areas and forests—across multiple mobile platforms. 
The setup employs stereo HD event cameras and records images, \mbox{LiDAR} and GPS data. 
However, despite their stated focus on high-speed dynamic motions in robotics applications, the large scene depths make the dataset essentially similar to DSEC rather than fundamentally different.

Our work is inspired by the UZH-FPV dataset~\cite{Delmerico19icra}, which captures UAV sequences under high-speed maneuvers using an mDAVIS event camera and evaluates sequence difficulty through optical flow distribution, rather than relying on absolute velocity. 
However, the monocular setup and relatively low resolution of this dataset limit its applicability in more diverse contexts. 
In contrast, our dataset incorporates a stereo event camera with VGA-resolution, extends the coverage to multiple widely used mobile robotic platforms, and provides ground truth data throughout via high-precision MoCap. 
Our dataset contributes to the advancement of event-camera-based state estimation in high-speed maneuvering scenarios. 

\section{Shortcomings of Current Datasets}
\label{sec: imperfections}
\begin{table}[t]
\centering
\caption{\label{tab:eventcount}\emph{Inconsistencies in stereo data.}
Comparison of event counts in stereo event camera datasets, with left-right differences ranging between $\approx$10\% ($\min$) and 85\% ($\max$). 
Sample event visualizations are given in Fig.~\ref{fig:stereo_comparison}.}
\begin{adjustbox}{max width=\linewidth}
\renewcommand{\arraystretch}{1.2}
\setlength{\tabcolsep}{2mm}{
\begin{tabular}{llrrr}
\toprule
\textbf{Dataset} & \textbf{Sequence} & \textbf{Left}  & \textbf{Right} & \textbf{Ratio} [\%] \\
\midrule
\multirow{2}{*}{MVSEC~\cite{Zhu18ral}} & \emph{indoor\_flying3} & 2.400$\times10^{7}$ & 2.188$\times10^{7}$ & +9.68  \\
{~} & \emph{indoor\_flying1} & 1.407$\times10^{7}$  & 1.200$\times10^{7}$ & +17.25 \\
\noalign{\vskip 2pt} \hline \noalign{\vskip 2pt}
\multirow{2}{*}{{HKU~\cite{Chen23ral}}} & \emph{hdr\_aggressive} & 1.285$\times10^{8}$  &1.435$\times10^{8}$  & -10.45 \\
{~} & \emph{aggressive\_small\_flip} & 1.031$\times10^{8}$  &1.516$\times10^{8}$  & -31.99 \\
\noalign{\vskip 2pt} \hline \noalign{\vskip 2pt}
\multirow{2}{*}{{DSEC}~\cite{Gehrig21ral}} & \emph{zurich\_city\_04b} & 1.290$\times10^{8}$ & 1.172$\times10^{8}$ & +10.07  \\
{~} & \emph{zurich\_city\_04f} & 3.615$\times10^{8}$ & 2.708$\times10^{8}$ & +33.49 \\
\noalign{\vskip 2pt} \hline \noalign{\vskip 2pt}
\multirow{2}{*}{{ECMD~\cite{Chen23ecmd}}} & \emph{street\_day\_difficult\_a2} &3.066$\times10^{7}$  & 2.253$\times10^{7}$  & +36.09  \\
{~} & \emph{street\_day\_easy\_a1} & 13.963$\times10^{7}$  &7.755$\times10^{7}$  & +85.82  \\
\bottomrule
\end{tabular}
}
\end{adjustbox}
\end{table}

\def\textWidth{0.035\linewidth}
\def\figWidth{0.312\linewidth} 
\begin{figure}[t]
	\centering
    {\scriptsize
    \setlength{\tabcolsep}{1pt}
	\begin{tabular}{
	>{\centering\arraybackslash}m{\textWidth}
	>{\centering\arraybackslash}m{\figWidth}
	>{\centering\arraybackslash}m{\figWidth}
    >{\centering\arraybackslash}m{\figWidth}}
    

        \rotatebox{90}{\makecell{aggressive\_small\_flip}}
		& \gframe{\includegraphics[width=\linewidth]{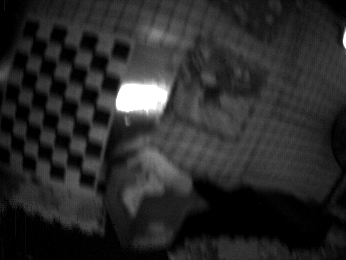}}
        & \gframe{\includegraphics[width=\linewidth]{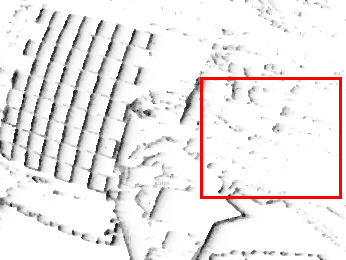}}
		& \gframe{\includegraphics[width=\linewidth]{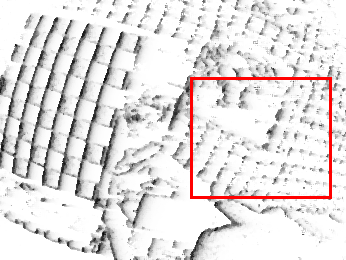}}
		\\   

        \rotatebox{90}{\makecell{zurich\_city\_04f}}
		& \gframe{\includegraphics[width=\linewidth]{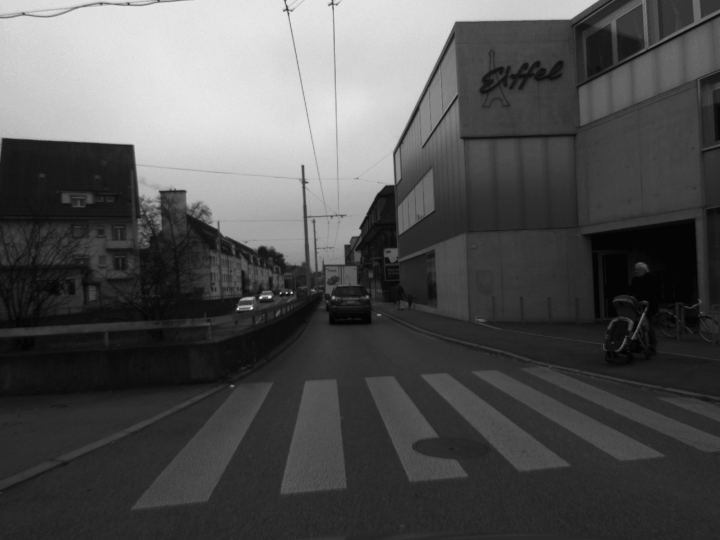}}
          & \gframe{\includegraphics[width=\linewidth]{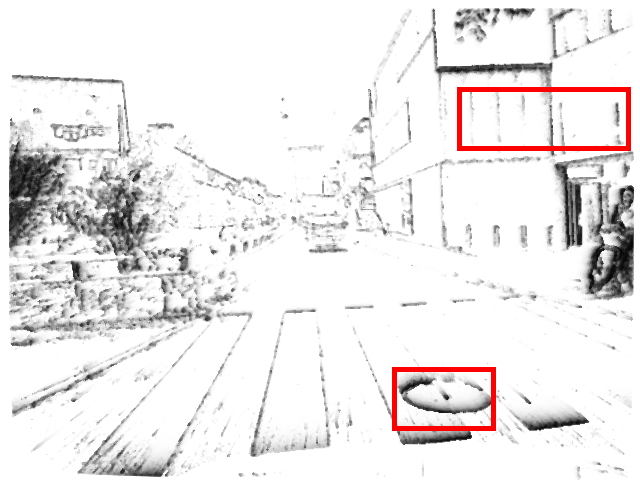}}
		& \gframe{\includegraphics[width=\linewidth]{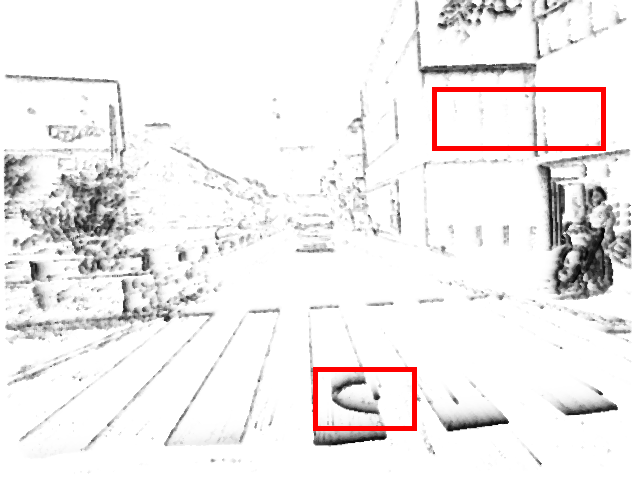}}
		\\   

        \rotatebox{90}{\makecell{street\_day\_easy\_a1}}
		& \gframe{\includegraphics[clip,trim={5.5cm 0 0 0},width=\linewidth]{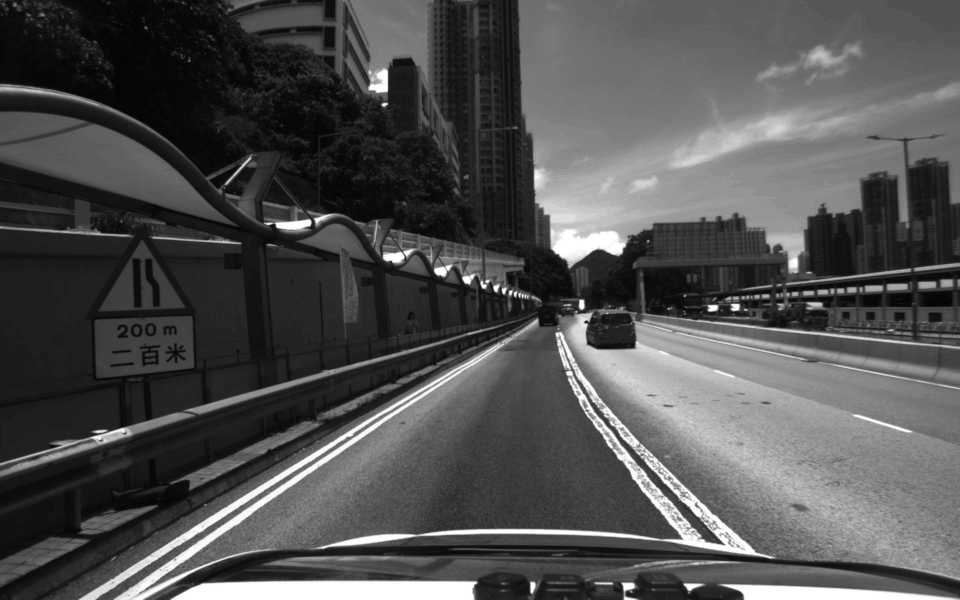}}
        & \gframe{\includegraphics[width=\linewidth]{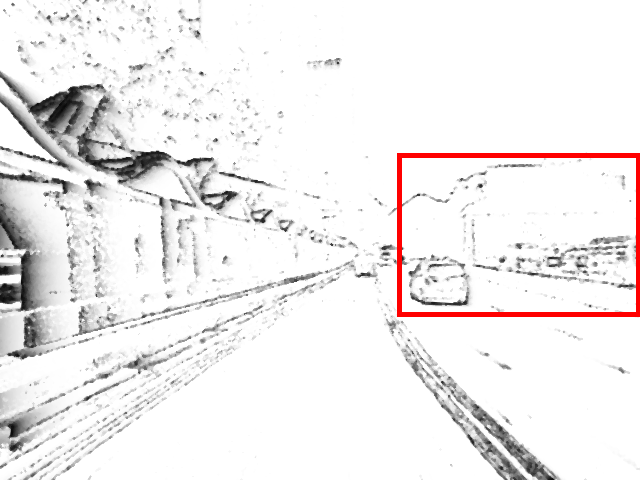}}
		& \gframe{\includegraphics[width=\linewidth]{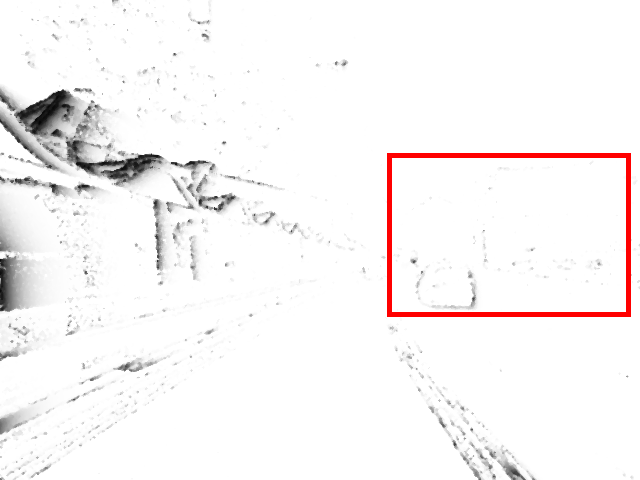}}
		\\   

		& Scene
        & Left event camera
        & Right event camera
		\\
	\end{tabular}
	}
	\caption{
    \emph{Inconsistencies in stereo data}.
    Comparison of left and right event time maps, 
    showing regions with particularly noticeable differences (red boxes).
    The corresponding images are provided for visualization purposes only.
    }
    \label{fig:stereo_comparison}
\end{figure}
Although numerous event-based SLAM datasets are available, most exhibit shortcomings that prevent them from fully evaluating the performance of event cameras in high-speed maneuvering scenarios. 
In this section, we combine theoretical analysis with experimental validation to discuss these shortcomings, including spatio-temporal inconsistency across stereo observations, the lack of comprehensive high-speed maneuvering scenarios, and the computational burden induced by excessively high spatial resolutions, and propose corresponding improvements.

\subsection{Spatio-Temporal Inconsistency in Stereo Observations}
It is well known that event data resemble the \emph{temporal} derivative of brightness \cite{Gallego20pami} and are, therefore, time-dependent by nature. 
Since motion depends on time, the events produced by the same camera and scene may differ (i.e., lack spatio-temporal consistency) depending on motion \cite{Gehrig19ijcv}.
A stereo configuration theoretically mitigates this issue, since the motion and FOV of the two cameras are highly similar.
However, we observe a substantial difference in event counts between the left and right cameras in several public datasets (Tab.~\ref{tab:eventcount}), with values ranging from 10\% to 86\%.
Considering the slight differences in FOV and motion patterns between the two cameras, count differences smaller than 10\% can be regarded as acceptable and are unlikely to have a noticeable impact given the inherent abundance and noise of event data.
Nevertheless, as differences increase, they become visually discernible directly in time maps~\cite{Lagorce17pami}, as shown in Fig.~\ref{fig:stereo_comparison}.
Many factors could contribute to this phenomenon, including variations in camera parameter settings and data blocking. 
Rather than enumerating all possibilities, we highlight a potentially overlooked critical factor: lens focus.

As noted in~\cite{Krotkov88ijcv}, defocus intrinsically acts as a low-pass filter that attenuates high-frequency information in an image.
In frame-based cameras, this attenuation manifests itself as a reduction of the intensity gradient, which serves as a metric to quantify the degree of defocus~\cite{Jahne05book}. 
For event cameras, \cite{Lin22cvpr} proposes the event generation rate as a quantitative indicator, based on the event generation model.
The underlying intuition is that defocus blurs scene edges on the image plane, making it more difficult for the pixel to reach the contrast threshold required to trigger events, thereby leading to a reduction of the event rate. 
This explanation is consistent with the observation in Fig.~\ref{fig:stereo_comparison}, where edges with low contrast are nearly invisible in one of the cameras.

We hereby propose a sensible focusing method applicable to stereo event cameras.
The cameras are directed toward a display screen while remaining stationary; 
a flickering checkerboard or a uniformly rotating line pattern is displayed on the screen, while the stereo event-count output per time unit (e.g., 100 ms) is monitored in real-time.
The focus rings of the each camera lens is continuously adjusted until the event rates reach their maximum and become closely matched, at which point the system is considered in focus.

\begin{figure}[t]
	\centering
    \includegraphics[width=0.8\linewidth]{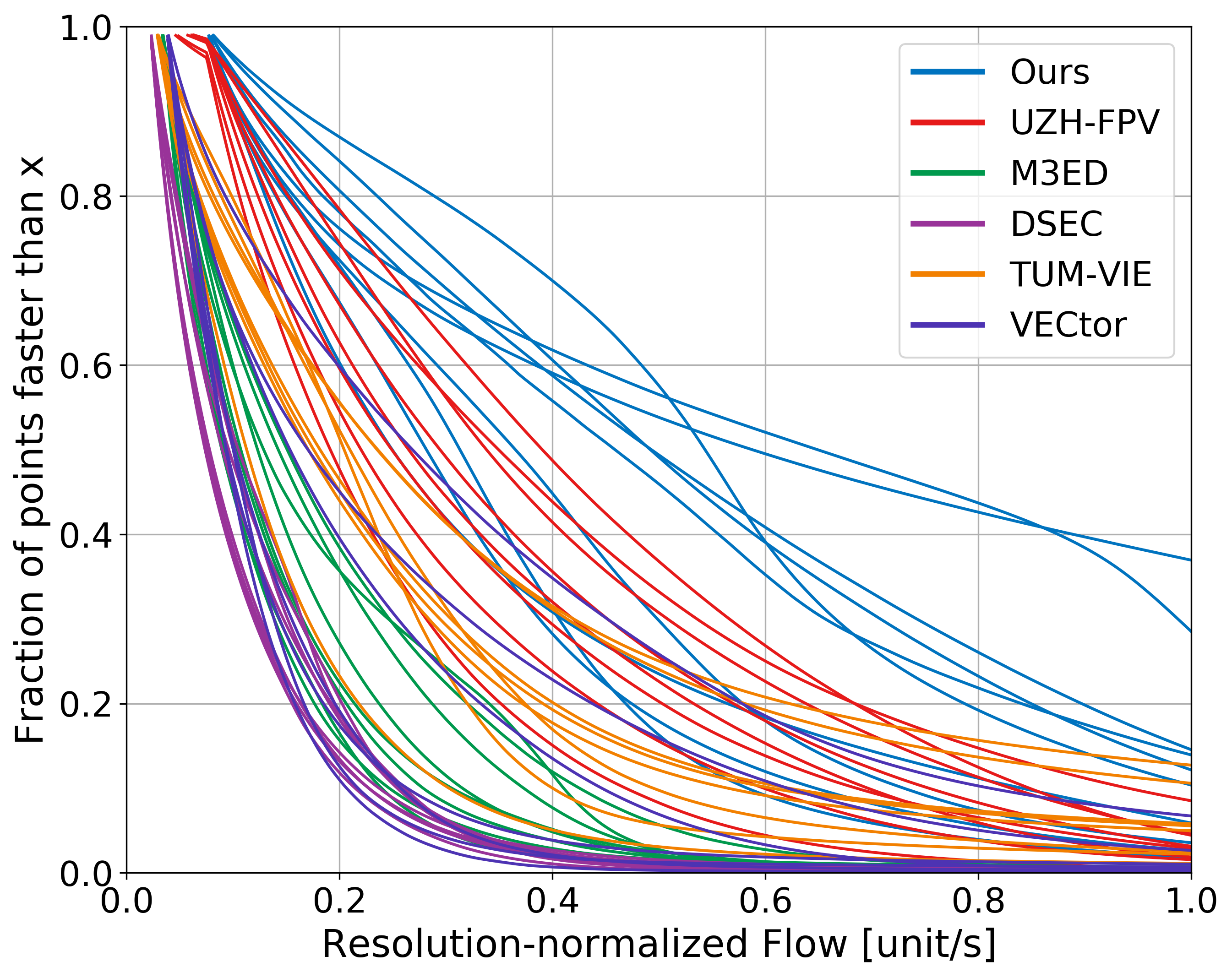}
	\caption{\label{fig:optical flow}\emph{Is existing data really from high-speed scenarios?}
    Comparison of resolution-normalized optical flow distributions for partial sequences across different datasets.} 
\end{figure}
\subsection{Scarcity of Data in High-Speed Maneuvering Scenarios}
\label{subsec: high-speed}

A definition of ``high-speed maneuvers'' must be explicitly formulated.
In our context, it should ($i$) address the primary causes of motion blur and ($ii$) focus on factors commonly encountered in practical applications of high-speed mobile robots. 
Specifically, our definition is based on two factors:
\begin{enumerate}
    \item A general 6-DoF motion pattern, where both linear and angular velocities can be sufficiently large to produce noticeable motion blur effects in the frames (if present);
    \item A proper motion parallax (i.e., a proper distance from the scene to the observer).
\end{enumerate}

Factor 1) distinguishes our defined scenarios from early works that typically demonstrated successful tracking of aggressive local shaking movements of an event camera~\cite{Gallego17pami,Gao22ral} (provided that the scene remained within the FOV). 
Such shaking patterns are, however, unlikely to occur in practical mobile robot applications.

Factor 2), inspired by \cite{Delmerico19icra}, indicates that motion blur is unlikely to occur when the scene depth is large, even at high linear velocities, while it can be induced when the observer is close to the scene, even with non-aggressive motion. 
From this perspective, we contend that employing image velocity (i.e., optical flow) rather than the camera's velocity, provides a more indicative criterion of the agility of the motion from the visual sensor's perspective. 
\textcolor{black}{Fig.~\ref{fig:optical flow} illustrates the distribution of optical flow (OF), normalized by the image diagonal, for selected sequences from multiple event-based datasets. 
All optical flow values are computed from the corresponding grayscale images using the same OpenCV function~\cite{Delmerico19icra}, with identical parameter settings across all sequences and extremely large and small flow magnitudes excluded.}
Although M3ED claims to focus on high-speed motions, its OF values are among the smallest, similar to DSEC, primarily due to the large scene depth. 
The other three datasets contain some sequences with relatively high OF magnitudes; however, these sequences suffer from limited scenario diversity. 
Specifically, VECtor and TUM-VIE predominantly consist of hand-held cameras being shaken in indoor scenes, while UZH-FPV contains only drone sequences.

In contrast, our dataset (Sec.~\ref{sec: dataset}) incorporates recordings from four distinct platforms, covering common high-speed maneuvering scenarios of mobile robots, thereby enabling more effective evaluation of algorithm performance under high-speed maneuvering scenarios.

\begin{figure}[t]
    \centering
    \includegraphics[width=\linewidth]{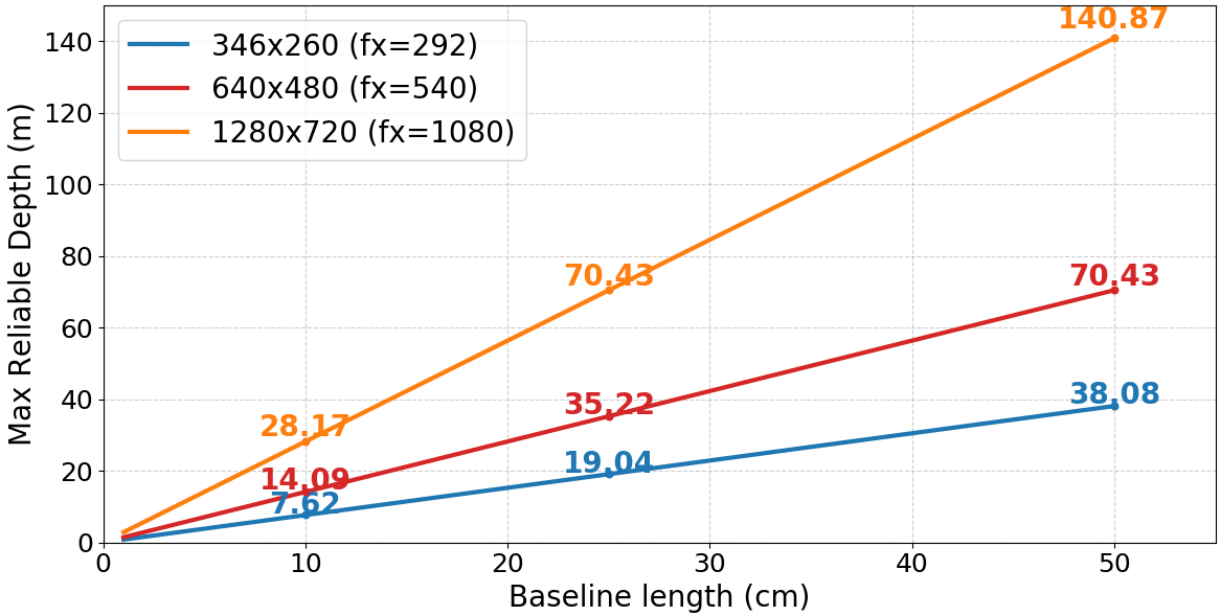}
	\caption{\textit{Maximum reliable depth estimation (with error~$\leq$15\%) for varying baselines and camera resolutions.} 
    The maximum depth values for 10 cm, 25 cm and 50 cm baselines are numerically highlighted.}
    \label{fig:maximum depth}
\end{figure}
\begin{figure*}[t]
	\centering

        

    \subfigure[Our setup]{
        \centering
        \gframe{\includegraphics[trim={250px 0 200px 0},clip,width=0.204\linewidth]{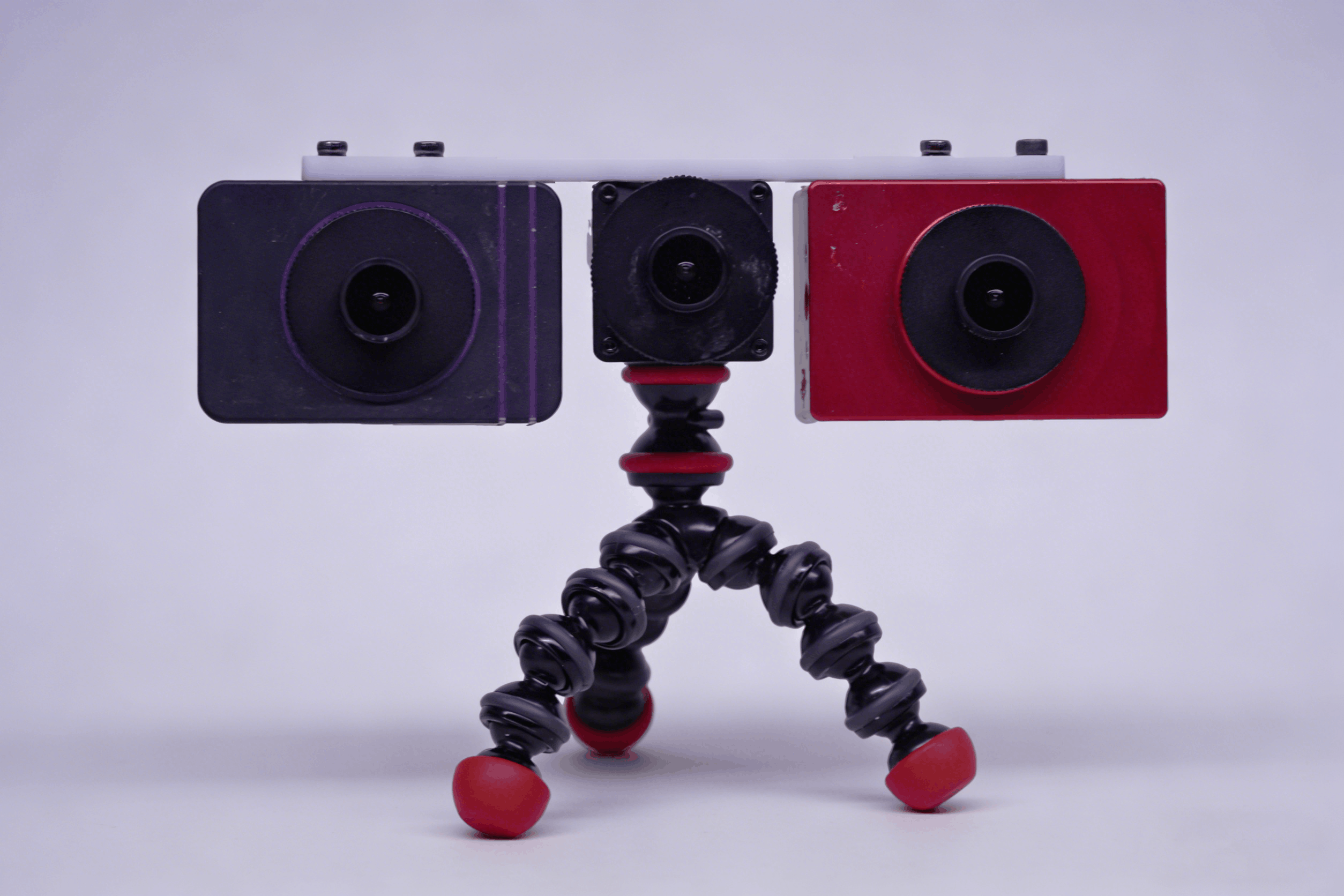}}
        \label{subfig:3_camera}
    }%
    \subfigure[\emph{DAVIS346} (346$\times$260 px)]{
        \centering
        \gframe{\includegraphics[width=0.22\linewidth]{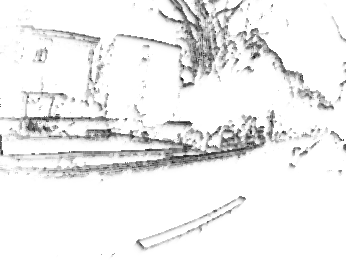}}
        \label{subfig:davis}
    }%
	\subfigure[\emph{DVXplorer} (640$\times$480 px)]{
        \centering
        \gframe{\includegraphics[width=0.22\linewidth]{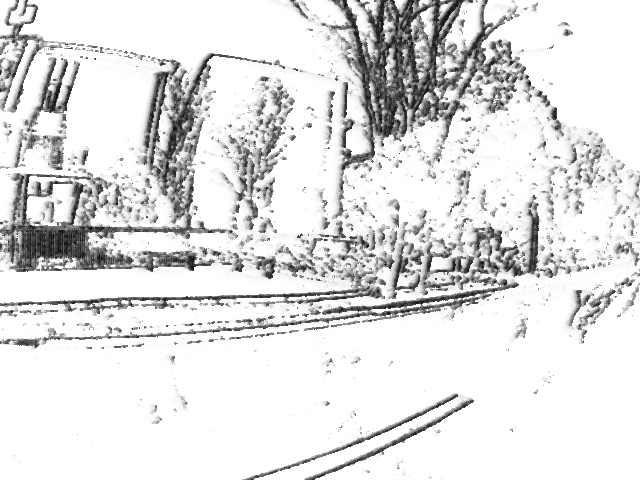}}
        \label{subfig:dvxplorer}
    }%
    \subfigure[\emph{Prophesee EVK4} (1280$\times$720 px)]{
        \centering
        \gframe{\imgbox{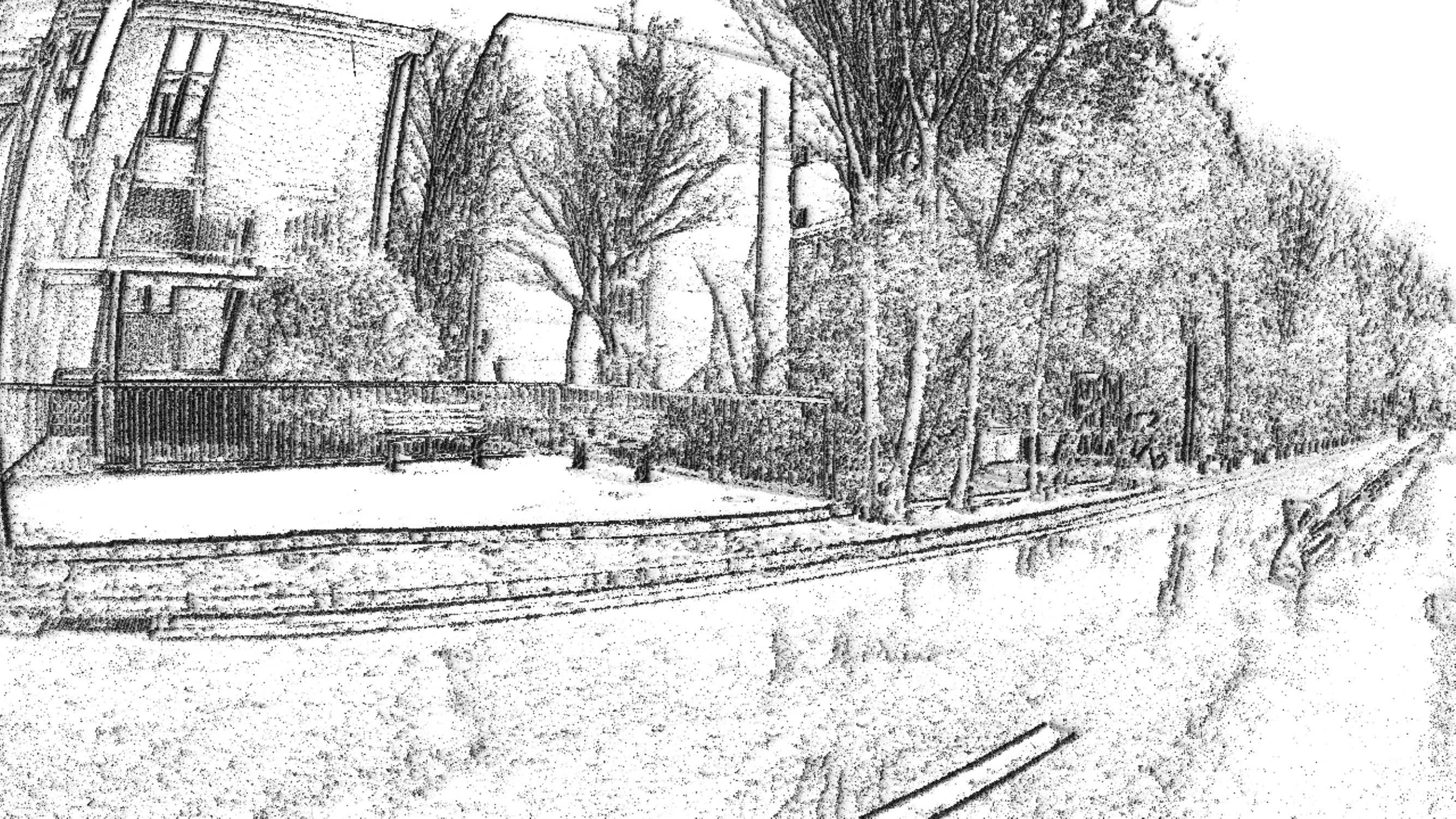}{(0.6,0.45) rectangle (0.85,0.9)}{0.292}}
        \label{subfig:prophesee}
    } 
	\caption{\emph{Effect of spatial resolution on event generation.}
    (a) depicts three side-by-side event cameras equipped with identical lenses and different resolutions, which are employed to record the data presented in Tab.~\ref{tab:event_statistics}. 
    (b)–(d) present the time maps captured by the setup shown in (a), sorted by increasing spatial resolution.
    }     
    \label{fig:mono_comparison}
\end{figure*}

\subsection{Unmanageable Data Rates due to the Megapixel Race}
\label{subsec: resolution}

Current mainstream event camera resolutions include 346$\times$260 px (low), 640$\times$480 px (middle), and 1280$\times$720 px (high) \cite{Gallego20pami,Qin25eventvisionsensorreview}. 
For event-based stereo VO/SLAM methods, increasing the spatial resolution has several effects on algorithmic performance:
($i$) it can improve the accuracy of stereo matching, 
leading to more precise depth estimation, 
($ii$) it enables the observation of more fine-scale scene details \cite{Ghosh22aisy}
and ($iii$) it can increase runtime (accuracy vs.~speed trade-off).

Regarding the first aspect, we calculate the maximum reliably estimated depth for various resolutions and baseline lengths (Fig.~\ref{fig:maximum depth}), with the criterion that the depth estimation error remains below 15\%. 
The maximum estimable depth $D_{\text{max}}$ is derived by imposing the relative error criterion:
$$
\frac{|D - \hat{D}|}{D} < 15\%,\quad D = f_x B/d,\quad \hat{D} = f_x B/(d+\Delta u)
$$
where $D$, $\hat{D}$ denote the estimated and true depth, respectively, $f_x$ the horizontal focal length, $B$ the stereo baseline, $d$ the true stereo disparity, and $\Delta u=\pm 0.5$ px the maximum disparity error under ideal conditions.
The results indicate that medium resolution with 10 cm and 25 cm baselines can adequately meet depth estimation requirements in most indoor and outdoor scenarios, respectively. 
The only exception is for drones in outdoor environments, as their stereo baseline is typically constrained to approximately 10 cm.

As for the second aspect, SLAM algorithms rely on event data triggered by stable, sharp edges to establish spatio-temporal correspondences. 
However, high-resolution sensors capturing excessive scene details often generate cluttered and non-discriminative events, called the ``Kirschbaum'' problem~\cite{Ponitz10aapr}, which interferes with this process. 
This interference is shown in Fig.~\ref{fig:mono_comparison}, which compares events at different resolutions under similar straight-driving scenarios. 
As clearly shown in Fig.~\ref{subfig:prophesee}, the cluttered events produced by fine textures obscure clear edges, thereby complicating the construction of reliable spatio-temporal correspondences.

\begin{table}[t]
\centering
\caption{\emph{Effect of spatial resolution on event generation}.
Event counts over five-second intervals for similar scenes with varying resolutions.
The number of events produced grows faster than the number of pixels.}
\label{tab:event_statistics}
\begin{adjustbox}{max width=\linewidth}

\begin{tabular}{llrr}
\toprule
\textbf{Dataset} & \textbf{Sequence} & \textbf{Resolution} [px]  & \textbf{Event Num} \\
\midrule
MVSEC & \emph{outdoor\_day\_2} & 346$\times$260 ($\sim$\hphantom{0}90$\times10^3$) & 3.265$\times10^6$ \\
DSEC & \emph{zurich\_city\_04a} & 640$\times$480 ($\sim$300$\times10^3$)& 24.267$\times10^6$\\
M3ED & \emph{car\_ucity\_small\_loop} & 1280$\times$720 ($\sim$900$\times10^3$) & 148.700$\times10^6$\\
\midrule
\multirow{3}{*}{Ours} & \emph{DAVIS346} & 346$\times$260 ($\sim$\hphantom{0}90$\times10^3$) & 1.202$\times10^6$ \\
{~} & \emph{DVXplorer} & 640$\times$480 ($\sim$300$\times10^3$)& 12.526$\times10^6$\\
{~} & \emph{Prophesee\_EVK4} & 1280$\times$720 ($\sim$900$\times10^3$) & 130.791$\times10^6$\\
\bottomrule
\end{tabular}
\end{adjustbox}
\end{table}
In terms of efficiency, the computational burden increases with resolution. 
Tab.~\ref{tab:event_statistics} reports event-count statistics on three datasets with different spatial resolutions, covering segments of the same duration and under similar driving scenarios. 
Additionally, we record an outdoor handheld sequence using three side-by-side cameras with identical lenses and different resolutions (the specific devices are illustrated in Fig.~\ref{subfig:3_camera}) under default driver parameters. 
The results demonstrate that as spatial resolution increases, each pixel becomes more sensitive to brightness changes~\cite{Gehrig22high}, resulting in a superlinear growth in event count with respect to spatial resolution.

The VO/SLAM task requires real-time performance, and current event-based SLAM methods~\cite{Niu25esvo2,Zhong25ral} can typically only operate in real-time at low or medium resolutions on high-end CPUs and GPUs.
\textcolor{black}{Hence, the event throughput at 1280$\times$720~px exceeds the above-mentioned processing capabilities of current methods. 
Although exploiting high-resolution event data remains a valuable direction for future research, we consider 640$\times$480~px the most suitable resolution for event-based SLAM at the present stage, given current processing capabilities and the intended use of our dataset.}

\section{Proposed Dataset}
\label{sec: dataset}
\subsection{Sensors and Platforms}
Driven by the requirements for high-speed maneuvers, our sensor suite comprises only a multi-camera setup, with inertial data provided by the built-in IMU of the event cameras. 
Further details regarding the characteristics of each sensor are provided in Tab.~\ref{tab:sensors}.
The stereo event camera has a resolution of 640$\times$480 px and a baseline of 10 cm. 
An Intel RealSense D435 is mounted above it, which incorporates two grayscale cameras with a 5 cm baseline and one color camera, all with a resolution of 640$\times$480 px. 
Unlike typical setups where infrared (IR)-blocking filters are used to mitigate interference from MoCap, we omit such filters in most sequences as both the D435 and our event camera are sensitive to IR light, which provides signal. 
Filters are employed solely during low-light and HDR scenarios to reduce overall illumination.
All sensors are rigidly attached via 3D-printed components to minimize overall weight.

To cover common high-speed maneuvering scenarios in mobile robotics, we collect data on four different platforms (Fig.~\ref{fig: eyecatcher}). 
Specifically, an FPV drone performs aggressive \mbox{6-DoF} motion, a linear slider produces pure translation, a robotic arm creates large-angle oscillations, and a Mecanum-wheeled robot performs rapid rotations.

\begin{table}[t]
  \centering
  \caption{Sensor specifications.}
  \label{tab:sensors}
  \renewcommand{\arraystretch}{1.2}
  \begin{tabularx}{\linewidth}{L{0.32\linewidth} Y}
    \toprule
    \textbf{Sensor Type} & \textbf{Description} \\
    \midrule
    \multirow{4}{*}{Event Cameras}  &$2\times$ Inivation DVXplorer\\ 
    {~}& $640\times480$ px, 10~cm baseline\\
    {~}&Monochrome, 110~dB dynamic range \\
    {~}&IMU: Bosch BMI-160, 6-axis, 800~Hz \\
    \midrule
    \multirow{4}{*}{Frame-based Cameras} & Intel RealSense D435\\ 
    {~}& 2$\times$ $640\times480$ px Grayscale, 5~cm baseline\\
    {~}&  1$\times$ $640\times480$ px RGB \\
    {~}& 30~Hz frame rate\\
    \bottomrule
  \end{tabularx}
\end{table}
\subsection{Calibration}

\begin{figure}[t]
	\centering
    {\includegraphics[width=0.9\linewidth]{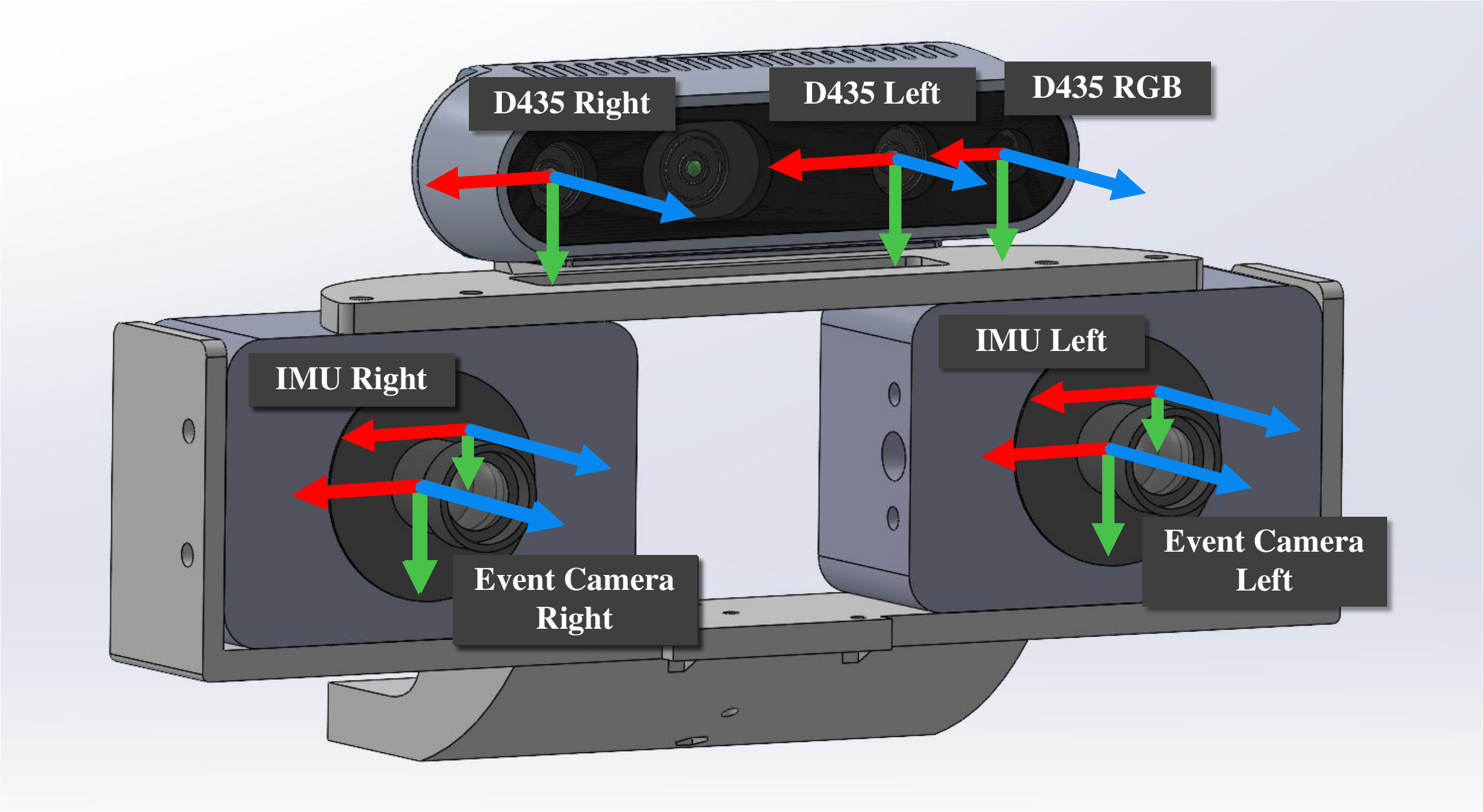}}
	\caption{\textit{Sensor suite CAD model with colored coordinate axes.}
    The axes of each sensor are: $X$ (red), $Y$ (green) and $Z$ (blue).} 
    \label{fig:sensor_suite}
\end{figure}

The CAD model of the sensor setup, featuring color-coded coordinate axes, is illustrated in Fig.~\ref{fig:sensor_suite}.
The calibration procedures for the respective intrinsic and extrinsic parameters of the vision sensors, IMU, and MoCap system are described as follows:

\subsubsection{Intrinsic and Extrinsic Calibration of Vision Sensors}
We estimate the intrinsic of event cameras and extrinsic parameters of the multi-camera setup independently using the Kalibr toolbox~\cite{Yan22opencalib}.
As Kalibr operates on image-based inputs, raw event streams are first converted into image frames using the Simple Image Reconstruction library \cite{Pfrommer23imgrec}. 
To ensure the precision of extrinsic parameter estimation, we first synchronize the event and RGB cameras via the synchronization scheme detailed in Sec.~\ref{subsec: time synchronization}, followed by image reconstruction conducted at the midpoint of each image exposure.

\subsubsection{Extrinsic Calibration between Vision Sensors and IMUs}
The spatiotemporal calibration of the vision sensors relative to the IMUs is performed by exciting the sensor suite with motion across all 6-DoF in front of an AprilTag grid. 
The recorded data sequences are then processed by the Kalibr toolbox~\cite{Yan22opencalib} to estimate the precise transformations.

\subsubsection{Hand-Eye Calibration}
To determine the extrinsic parameters between the MoCap system and the vision sensors, we use an AprilTag calibration board. 
The camera captures the AprilTag, while the MoCap system records its pose synchronously. 
The camera pose relative to the AprilTag is first estimated via the PnP algorithm~\cite{Li12pami}. 
These poses and those recorded by the MoCap system are passed as input to a hand–eye calibration method that estimates the desired rigid-body transformation~\cite{Daniilidis99ijrr}.

\begin{table*}[t]
\centering
\caption{Experimental setup and characteristics of each sequence.}
\label{tab: sequences}
\begin{adjustbox}{max width=\linewidth}
\begin{tabular}{l l c c c c l} 
\toprule
\textbf{Sequence Name} & \textbf{Setup} & \textbf{Duration} [s] & $\| \mathbf{v}\|_{\max}$ [m/s] & $\| \boldsymbol{\omega}\|_{\max}$ [rad/s] & \textbf{AUC of OF} & \textbf{Motion Pattern} \\ 
\midrule
\textit{drone\_circle\_norm} & 250mm quadcopter & 41.0& 6.93  & 1.18   & 0.307 & Saddle-circular motion \\
\textit{drone\_circle\_fast} & 250mm quadcopter & 38.9& 8.64  & 1.08  & 0.386 & Saddle-circular motion \\
\textit{drone\_line\_norm} & 250mm quadcopter & 41.2& 3.10  & 1.67  & 0.189 & Linear motion (w/o\ yaw) \\
\textit{drone\_line\_fast} & 250mm quadcopter & 33.4& 4.66  & 0.47  & 0.249 & Linear motion (w/o\ yaw) \\
\textit{drone\_8\_norm} & 250mm quadcopter & 41.4& 7.26  & 1.68  & 0.269 & Lemniscate trajectory motion \\
\textit{drone\_8\_hdr} & 250mm quadcopter & 39.9& 7.20  & 1.74  & 0.263 & Lemniscate trajectory motion \\
\textit{drone\_s\_norm} & 250mm quadcopter & 47.7& 4.07  & 2.22  & 0.292 & Sinusoidal trajectory motion \\
\textit{drone\_s\_hdr} & 250mm quadcopter & 45.0& 4.01  & 2.21  & 0.272 & Sinusoidal trajectory motion \\
\textit{drone\_rot\_norm} & 250mm quadcopter & 24.6& 1.07  & 2.19  & 0.590 & Circling with self-rotation \\
\textit{drone\_rot\_fast} & 250mm quadcopter & 24.9& 1.03  & 3.28  & 0.721 & Circling with self-rotation \\
\textit{slider\_norm} & Rail-mounted platform & 12.8& 1.60  & 0.14  & 0.293 & Linear motion (w/o\ yaw) \\
\textit{slider\_fast} & Rail-mounted platform & 7.1& 2.27  & 0.53  & 0.480 & Linear motion (w/o\ yaw) \\
\textit{slider\_hdr} & Rail-mounted platform & 5.9& 2.22  & 0.11  & 0.447 & Linear motion (w/o\ yaw) \\
\textit{arm\_norm} & Robotic arm & 26.8& 1.78  & 3.57  & 0.489 & Significant camera shake \\
\textit{arm\_dim} & Robotic arm & 22.1& 1.69  & 3.54  & 0.501 & Significant camera shake, \\
\textit{arm\_hdr} & Robotic arm & 26.4& 1.72  & 3.61  & 0.478 & Significant camera shake \\
\textit{car\_circle\_norm} & Mecanum-wheeled robot & 35.9& 2.28  & 1.39  & 0.504 & Circling \\
\textit{car\_circle\_hdr} & Mecanum-wheeled robot & 21.0& 2.04  & 1.27  & 0.402 & Circling \\
\textit{test} & Mecanum-wheeled robot & 22.0 & 2.15  & 1.39  & 0.386 & Circling \\
\bottomrule
\end{tabular}
\end{adjustbox}
\end{table*}


\subsection{Time Synchronization}
\label{subsec: time synchronization}
To achieve temporal synchronization between the DVXplorer event cameras and the D435 camera, a global clock synchronization scheme based on external hardware triggering is implemented~\cite{Gehrig21ral}. 
A pulse generator delivers synchronization pulses simultaneously to both event cameras and the D435 camera at a frequency of 30 Hz. 
Upon receiving each pulse, the event cameras insert a dedicated marker into their data stream at the corresponding timestamp, while the D435 camera triggers an image capture at each pulse.

During the data post-processing stage, the recorded markers in the event camera data are aligned with the timestamps of the images captured by the D435, establishing precise temporal correspondence across modalities. 
This hardware synchronization method ensures high-precision time alignment within the multi-sensor system and provides a reliable foundation for subsequent cross-modal data analysis.


\subsection{Sequences Overview}
The basic characteristics of all sequences are summarized in Tab.~\ref{tab: sequences}. 
Each sequence's name indicates the platform used, while the suffix denotes the motion speed or complex lighting condition, with drone sequences further specified by their motion patterns.
As mentioned in Sec.~\ref{subsec: high-speed}, the maximum linear and angular velocities do not adequately measure the level of high-speed maneuvering of a sequence.
Therefore, we also use the area under the resolution-normalized optical flow distribution curve~\cite{Delmerico19icra} as an intuitive and rough metric to compare the difficulty of different sequences, which we refer to as the ``AUC of OF''.
All sequences are recorded within the observable boundaries of the MoCap system and incorporate multiple closed loops, whose ground-truth trajectories are derived by fusing hand-eye calibration results with MoCap data, yielding sub-millimeter-level accuracy for 6-DoF pose estimation. 
Specifically, the LUSTER FZMotion MoCap system, operating at a 120 Hz sampling rate, is deployed for drone sequences, while the NOKOV Mars26H system with a 60 Hz sampling rate is used for all remaining.
\textcolor{black}{Regarding event camera settings, the DVXplorer provides five sensitivity levels, ranging from 0 to 4. 
Higher sensitivity levels correspond to lower event-triggering thresholds, allowing smaller brightness changes to generate events. 
We use the default level of 2 under normal illumination and increase it to 3 in HDR or low-light scenes to capture more events.}
Visualizations of sample sequences can be found in Fig.~\ref{fig: sequences_overview}.

Due to the high-speed maneuvers, the drone sequences are recorded separately in a large-scale flight arena that allows natural light to pass through.
The arena, enclosed by safety nets, experiences slight light interference, and striped boxes are placed within it.
The drone sequences comprise five motion patterns with progressively increasing difficulty: linear reciprocating motion, elliptical motion, figure-eight motion, \mbox{S-curve} motion, and circular motion with self-rotation, covering common high-speed maneuvering scenarios.
In contrast, the remaining sequences are recorded in a small-scale flight field with abundant texture information, where rail-mounted platform, robotic arm, and mecanum-wheeled robot perform linear reciprocating motion, back-and-forth shaking, and circular motion, respectively.

\def\textWidth{0.02\linewidth}
\def\figWidth{0.15\linewidth} 
\begin{figure*}[t]
	\centering
    {\scriptsize
    \setlength{\tabcolsep}{1pt}
	\begin{tabular}{
	>{\centering\arraybackslash}m{\textWidth}
	>{\centering\arraybackslash}m{\figWidth}
    >{\centering\arraybackslash}m{\figWidth}
    >{\centering\arraybackslash}m{\figWidth}
    >{\centering\arraybackslash}m{\textWidth}
    >{\centering\arraybackslash}m{\figWidth}
    >{\centering\arraybackslash}m{\figWidth}
	>{\centering\arraybackslash}m{\figWidth}}
    
        \rotatebox{90}{\makecell{drone\_circle\_fast}}
		& \gframe{\includegraphics[width=\linewidth]{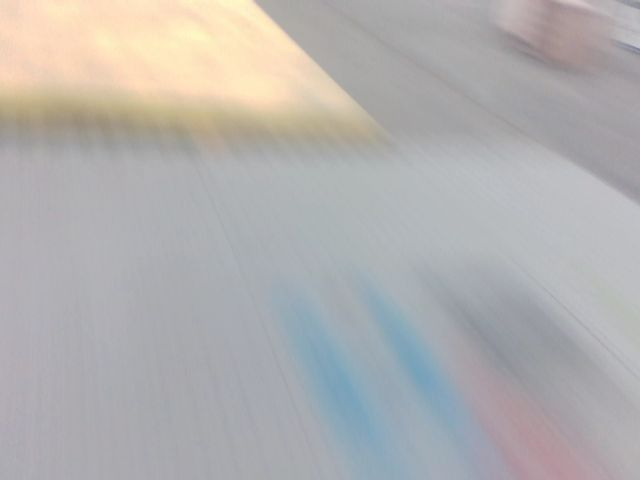}}
		& \gframe{\includegraphics[width=\linewidth]{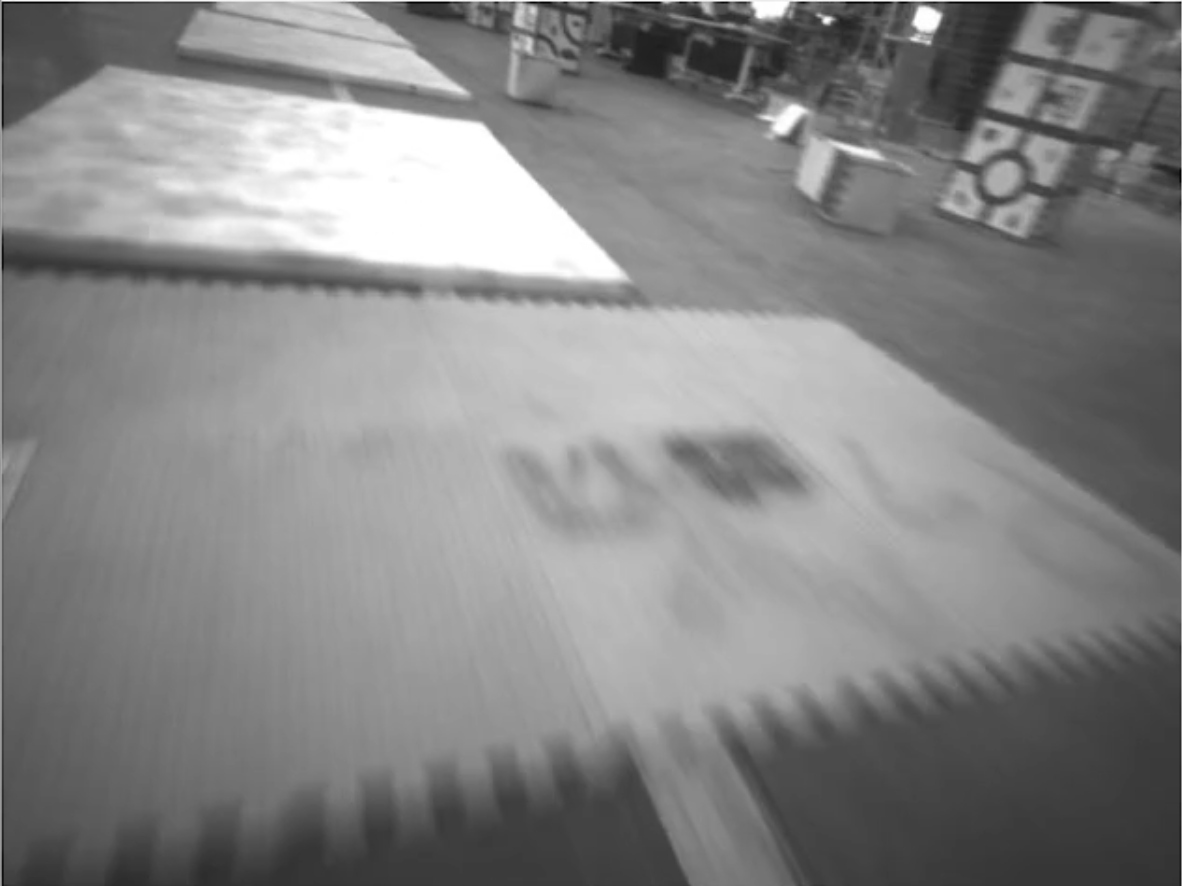}}
		& \gframe{\includegraphics[width=\linewidth]{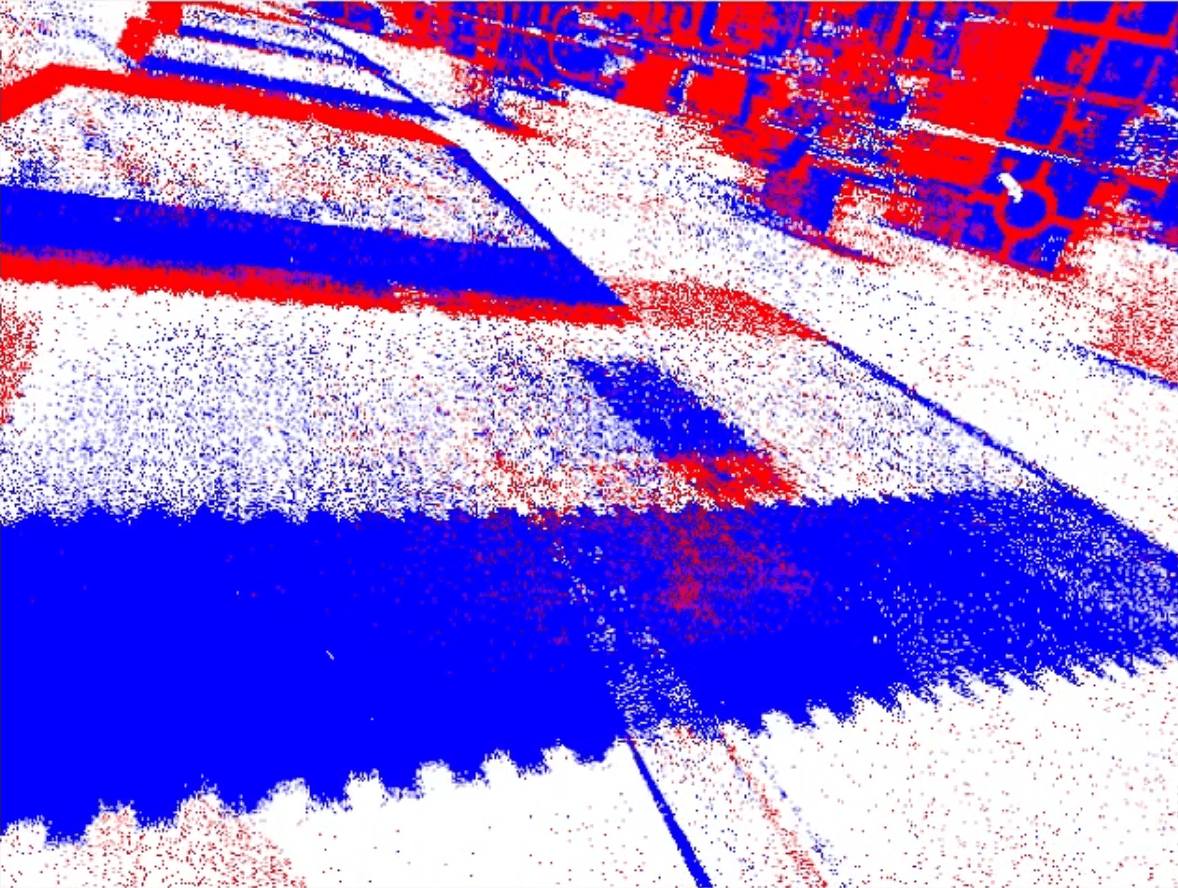}}
        & \rotatebox{90}{\makecell{drone\_8\_norm}}
        & \gframe{\includegraphics[width=\linewidth]{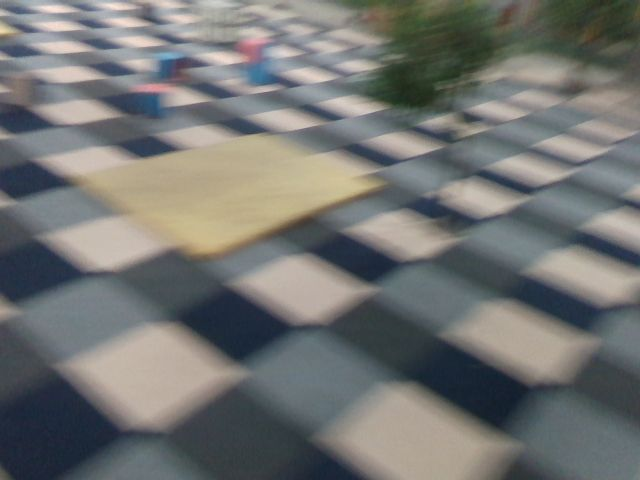}}
		& \gframe{\includegraphics[width=\linewidth]{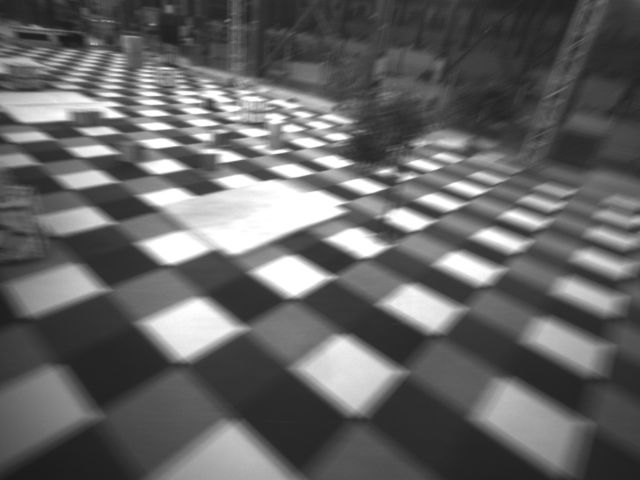}}
		& \gframe{\includegraphics[width=\linewidth]{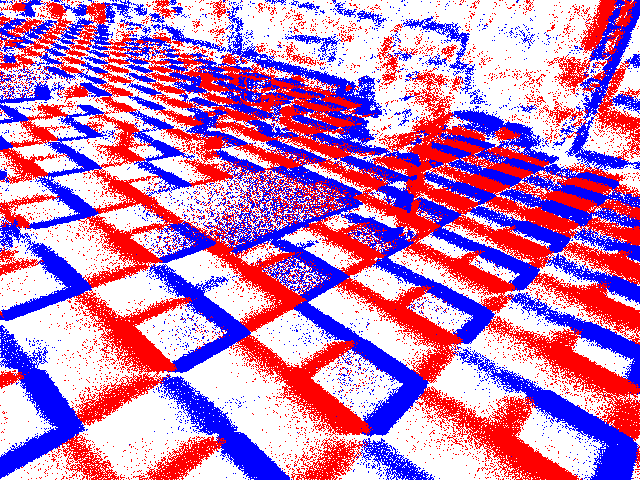}}
		\\   

        \rotatebox{90}{\makecell{arm\_norm}}
		& \gframe{\includegraphics[width=\linewidth]{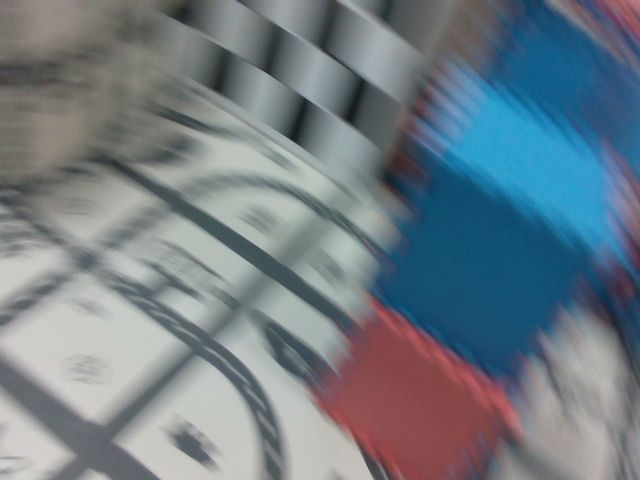}}
		& \gframe{\includegraphics[width=\linewidth]{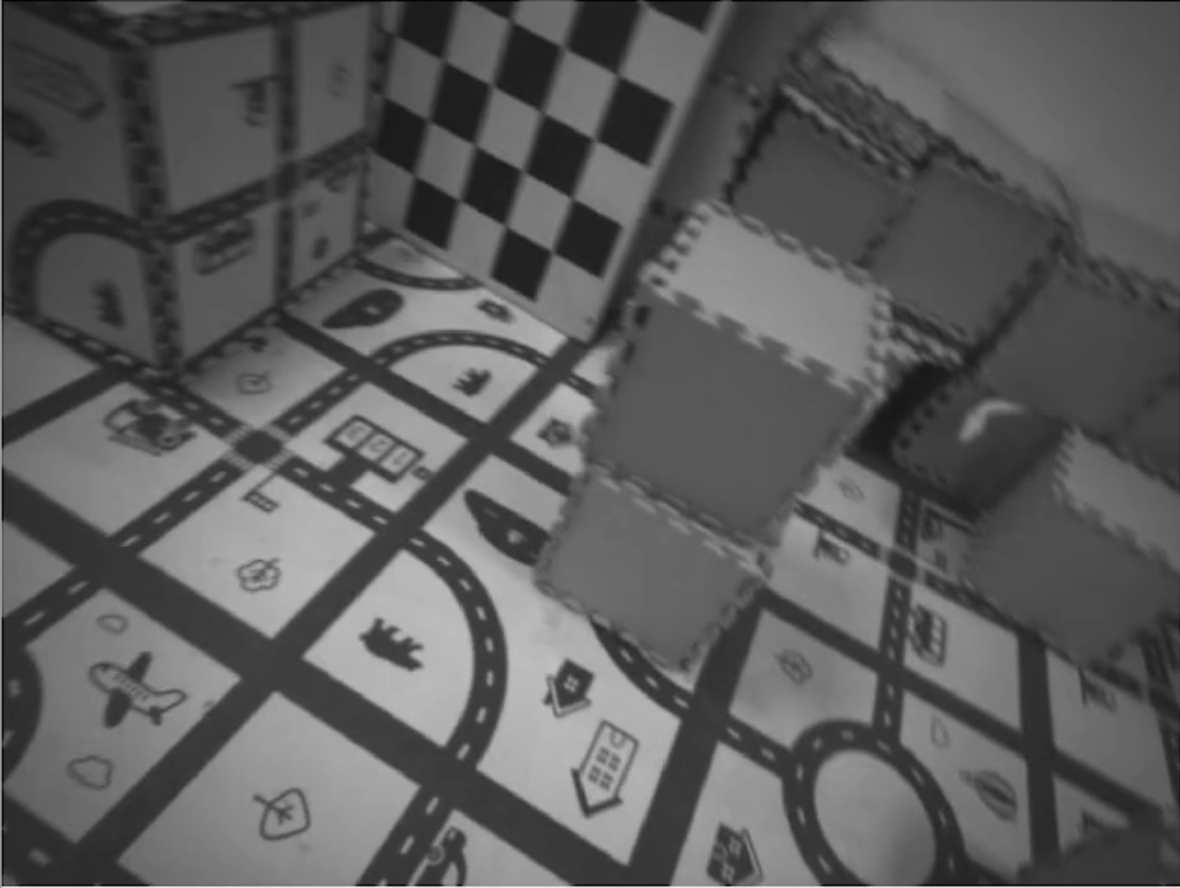}}
		& \gframe{\includegraphics[width=\linewidth]{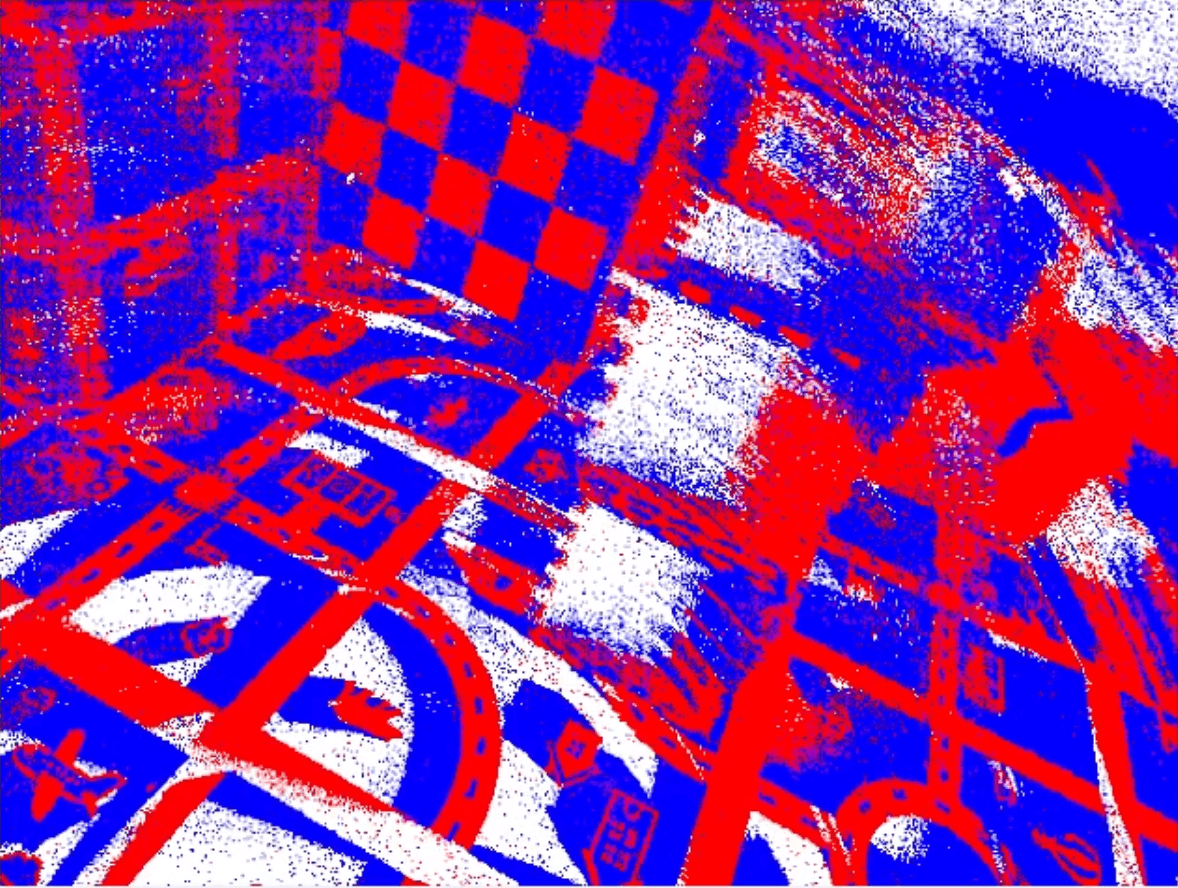}}
        & \rotatebox{90}{\makecell{drone\_8\_hdr}}
        & \gframe{\includegraphics[width=\linewidth]{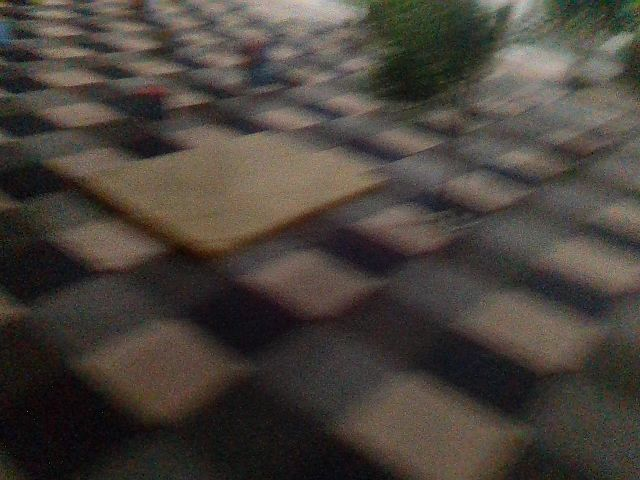}}
		& \gframe{\includegraphics[width=\linewidth]{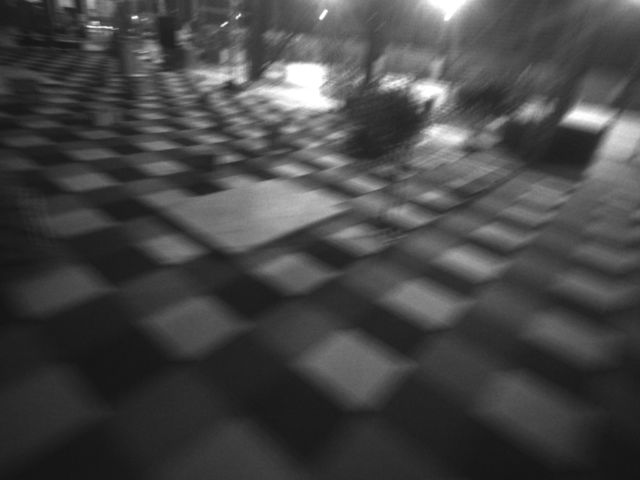}}
		& \gframe{\includegraphics[width=\linewidth]{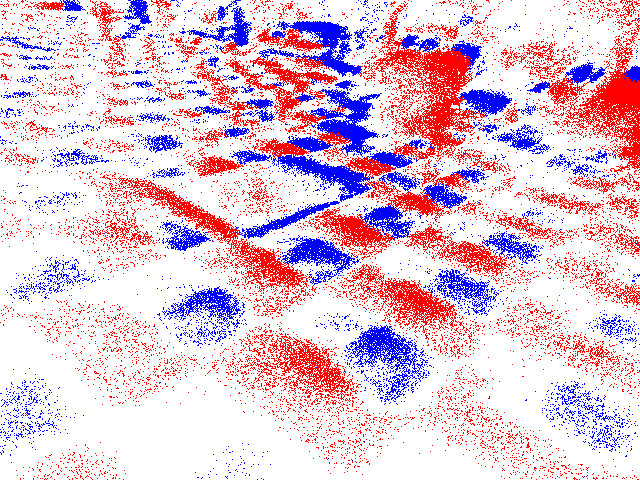}}
		\\   
        
        \rotatebox{90}{\makecell{drone\_s\_hdr}}
		& \gframe{\includegraphics[width=\linewidth]{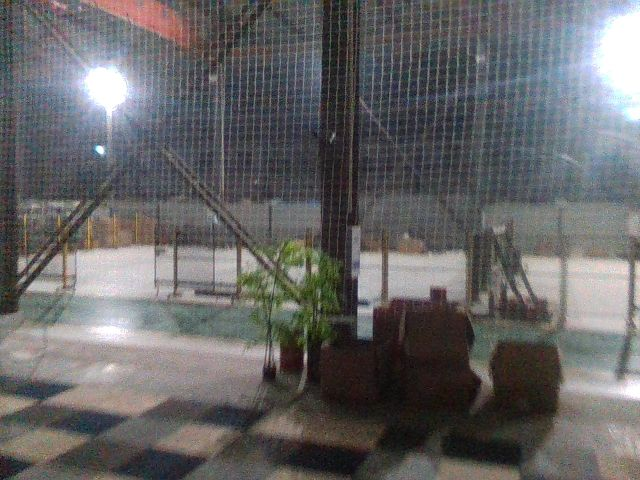}}
		& \gframe{\includegraphics[width=\linewidth]{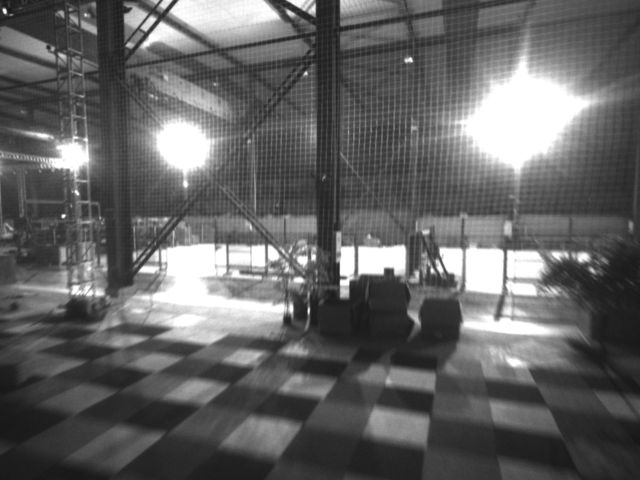}}
		& \gframe{\includegraphics[width=\linewidth]{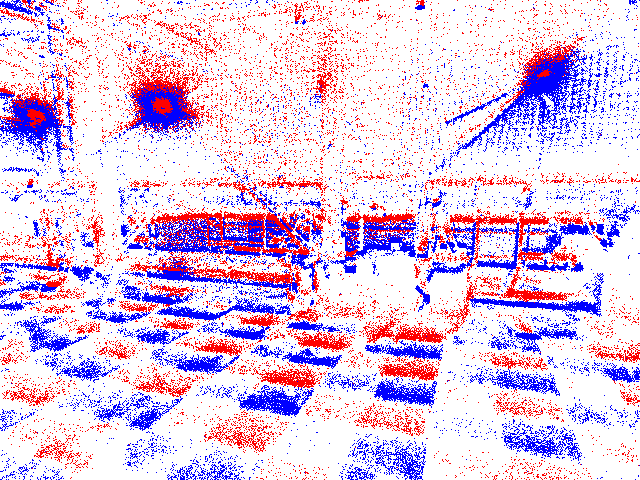}}
        & \rotatebox{90}{\makecell{arm\_hdr}}
        & \gframe{\includegraphics[width=\linewidth]{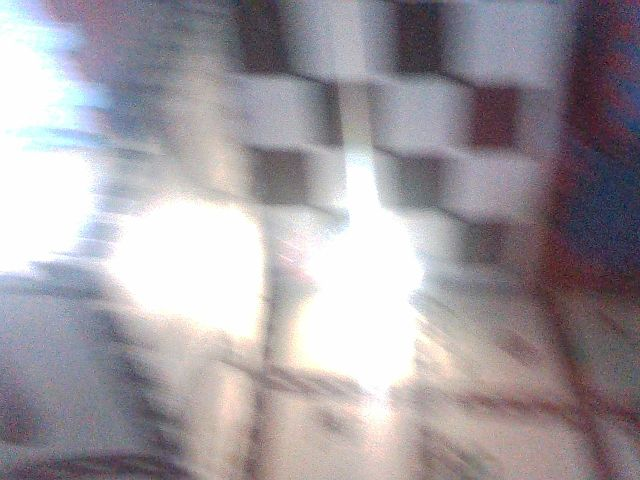}}
		& \gframe{\includegraphics[width=\linewidth]{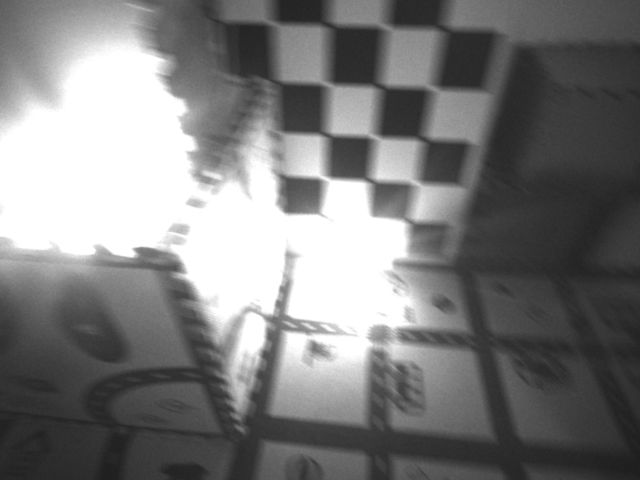}}
		& \gframe{\includegraphics[width=\linewidth]{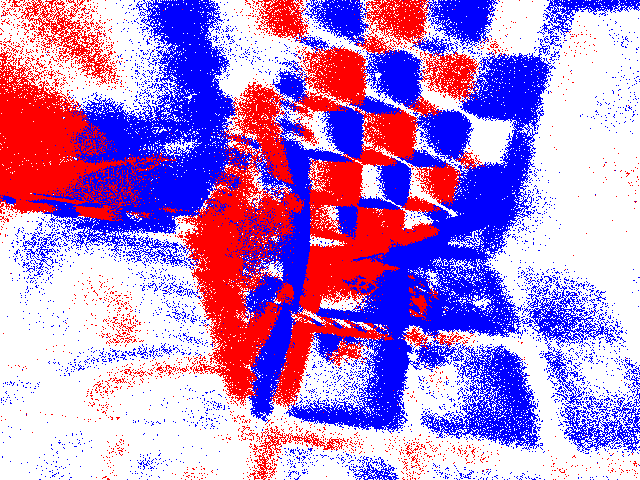}}
		\\

        \rotatebox{90}{\makecell{slider\_hdr}}
		& \gframe{\includegraphics[width=\linewidth]{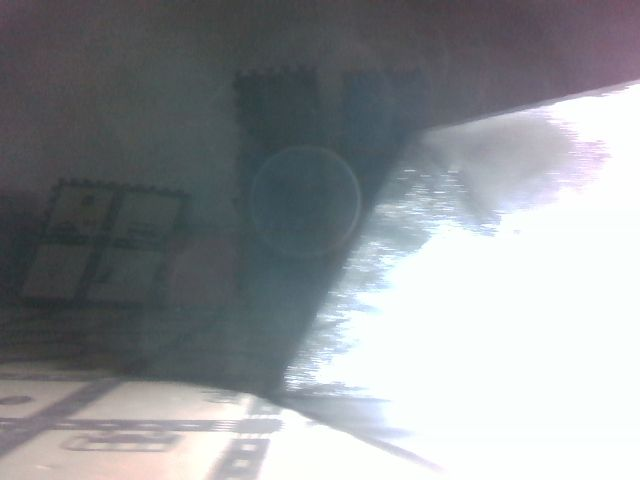}}
		& \gframe{\includegraphics[width=\linewidth]{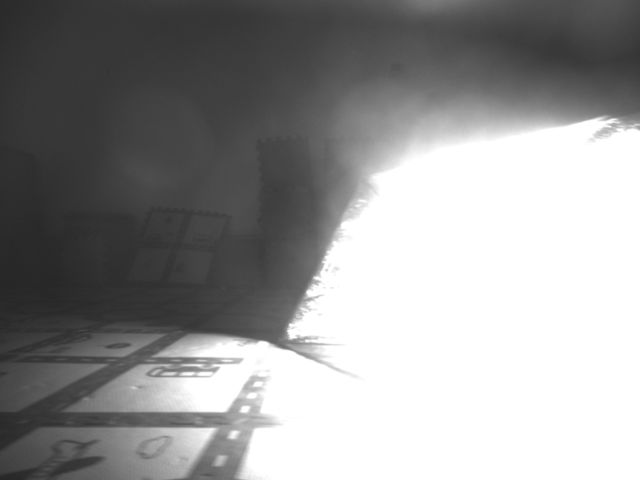}}
		& \gframe{\includegraphics[width=\linewidth]{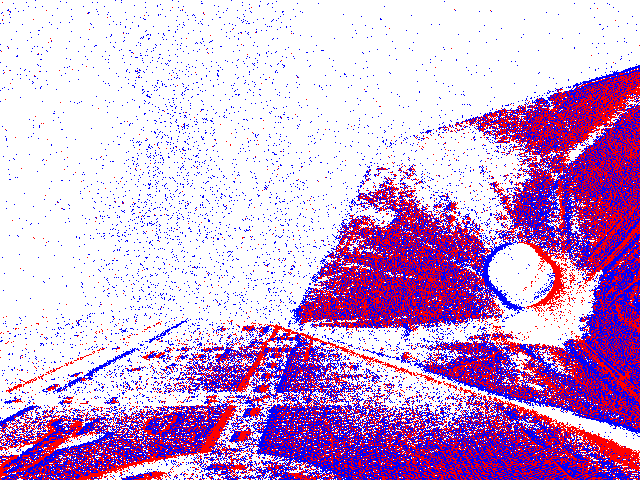}}
        & \rotatebox{90}{\makecell{car\_circle\_hdr}}
        & \gframe{\includegraphics[width=\linewidth]{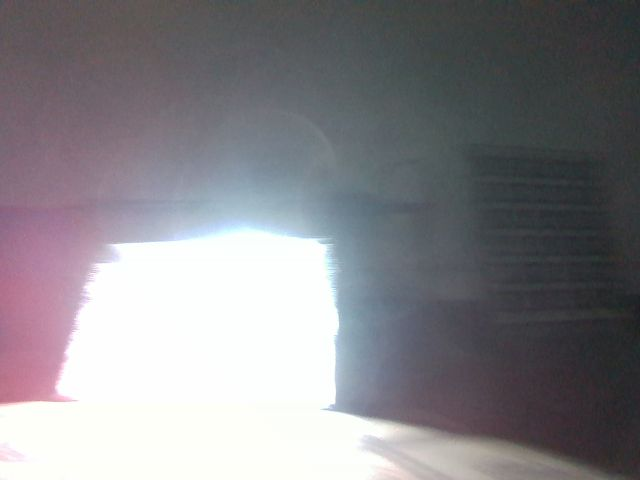}}
		& \gframe{\includegraphics[width=\linewidth]{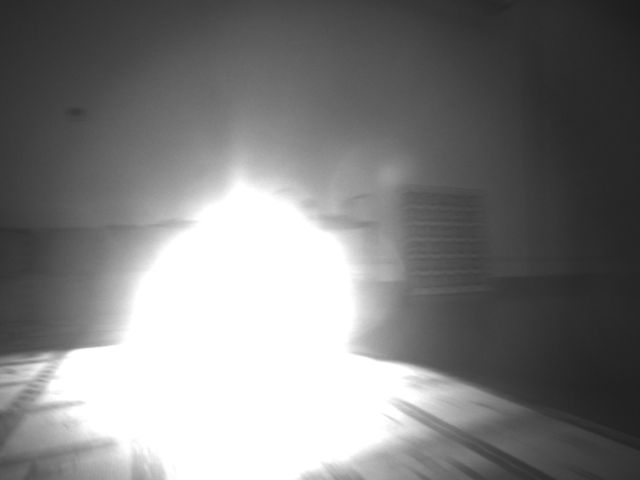}}
		& \gframe{\includegraphics[width=\linewidth]{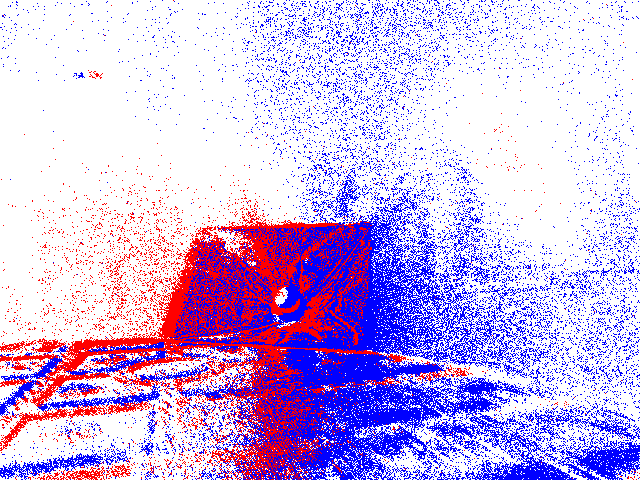}}
		\\
        
		& RGB
        & Grayscale
        & Events
        &
        & RGB
        & Grayscale
        & Events
		\\
	\end{tabular}
	}
    \caption{Visualization of sample sequences of the EvSLAM dataset.}
    \label{fig: sequences_overview}
\end{figure*}

Data collection is conducted in two batches, with the first batch including the first four drone sequences and the first sequence from each of the other platforms.
Although these sequences already exhibit significant optical flow and noticeable motion blur in the color images, grayscale images, benefiting from higher light intake and shorter exposure times, suffer from little motion blur for the majority of the time across most sequences, except the robotic arm ones, allowing frame‑based methods to perform well.
Therefore, in the second batch, we further increase the angular velocity and incorporate HDR and low-light scenarios to induce more pronounced motion blur in grayscale images, thereby better demonstrating the potential of event cameras in handling extremely challenging scenarios.
Moreover, since the drone sequences in the second batch are recorded during the evening rather than in the afternoon, the overall illumination is lower than that of the first batch.

\section{Benchmark}
\label{sec: benchmark}

\subsection{Evaluation Metrics}

The state variables of a VIO pipeline commonly include $\{\mathbf{p}, \mathbf{q}, \mathbf{v}, \mathbf{b}_{\text{a}}, \mathbf{b}_{\text{g}}\}$, namely the position $\mathbf{p}$, orientation $\mathbf{q}$, linear velocity $\mathbf{v}$, and sensor biases $\{\mathbf{b}_{\text{a}}, \mathbf{b}_{\text{g}}\}$ of the inertial measurement unit (IMU). 
The most widely used evaluation metrics for VIO systems are Absolute Trajectory Error (ATE) and Relative Pose Error (RPE)~\cite{Sturm12iros}.
Let the errors at timestamp $i$ between ground truth and estimated poses $\mathbf{T}$ be
\begin{equation}
\begin{split}
\text{ATE}_i &= \big\| \log\big( \mathbf{T}_{\text{gt},i}^{-1} \mathbf{T}_{\text{est},i} \big)^\vee \big\|_2 , \\
\text{RPE}_i &= \big\| \log\big( ( \mathbf{T}_{\text{gt},i}^{-1} \mathbf{T}_{\text{gt},i+\Delta} )^{-1} 
                                    ( \mathbf{T}_{\text{est},i}^{-1} \mathbf{T}_{\text{est},i+\Delta} ) \big)^\vee \big\|_2,
\end{split}
\label{eq:ate and rpe}
\end{equation}
it is common to define the ATE and RPE in terms of the root mean square (RMS) values over the whole trajectory:
\begin{equation}
\text{ATE} \doteq \frac1{n} \sqrt{\sum_{i=1}^{n} \text{ATE}_i^2}, \quad
\text{RPE} \doteq \frac1{n-\Delta} \sqrt{\sum_{i=1}^{n-\Delta} \text{RPE}_i^2}.
\label{eq:ATE-RPE:RMS}
\end{equation}

The ATE measures the global consistency of the estimated trajectory with respect to the ground truth, making it particularly useful for assessing long-term drift and overall accuracy.
The RPE, on the other hand, evaluates short-term tracking performance by analyzing pose errors over fixed time intervals or displacements (of size $\Delta$). 
It is commonly expressed in terms of drift per second or per frame, providing insight into local consistency and incremental motion estimation.
While both metrics are effective for assessing pose accuracy, instantaneous linear velocity estimation remains an under-examined aspect in VIO performance assessment.

\begin{figure*}[t!]
  \centering
  \captionsetup{skip=1ex}

  \subfigure[\small{RVE statistics (-).}]{\raisebox{1cm}{
    \noindent{\scriptsize
    \begin{tabular*}{0.25\textwidth}{lccc}
    \toprule
    Method & Mean & Median & Std Var\\ 
    \midrule
    ESIO & 0.226 & 0.168 & 0.222 \\
    EVIV & 0.132 & 0.111 & 0.087 \\
    USLAM & 0.154 & 0.144 & 0.068 \\
    \bottomrule
    \end{tabular*}
   }
  }
  }~
  \subfigure[\small{RVE boxplot statistics.}]{
  \includegraphics[width=0.20\textwidth]{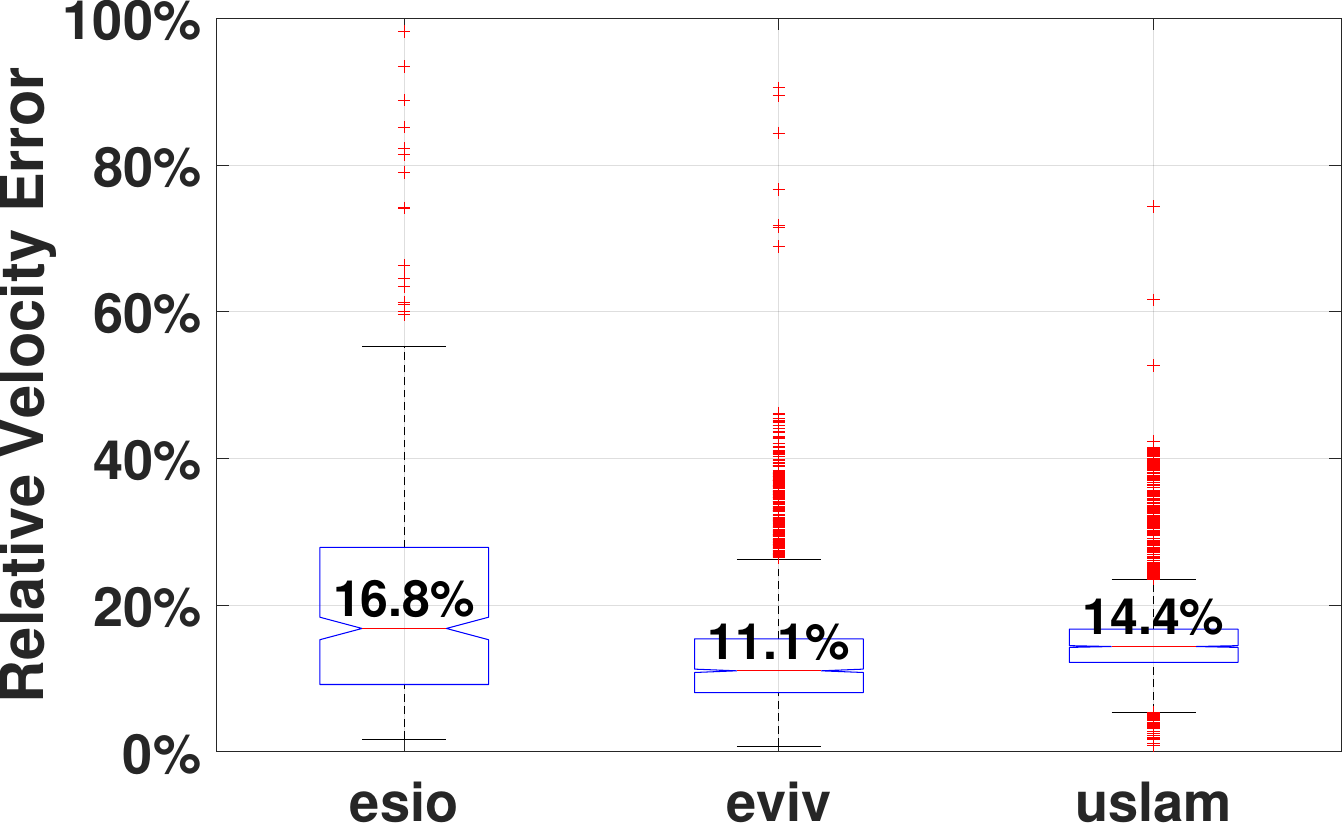}}~
  \subfigure[\small{Unweighted $S({\xi})$ curves.}]{
  \includegraphics[width=0.20\textwidth]{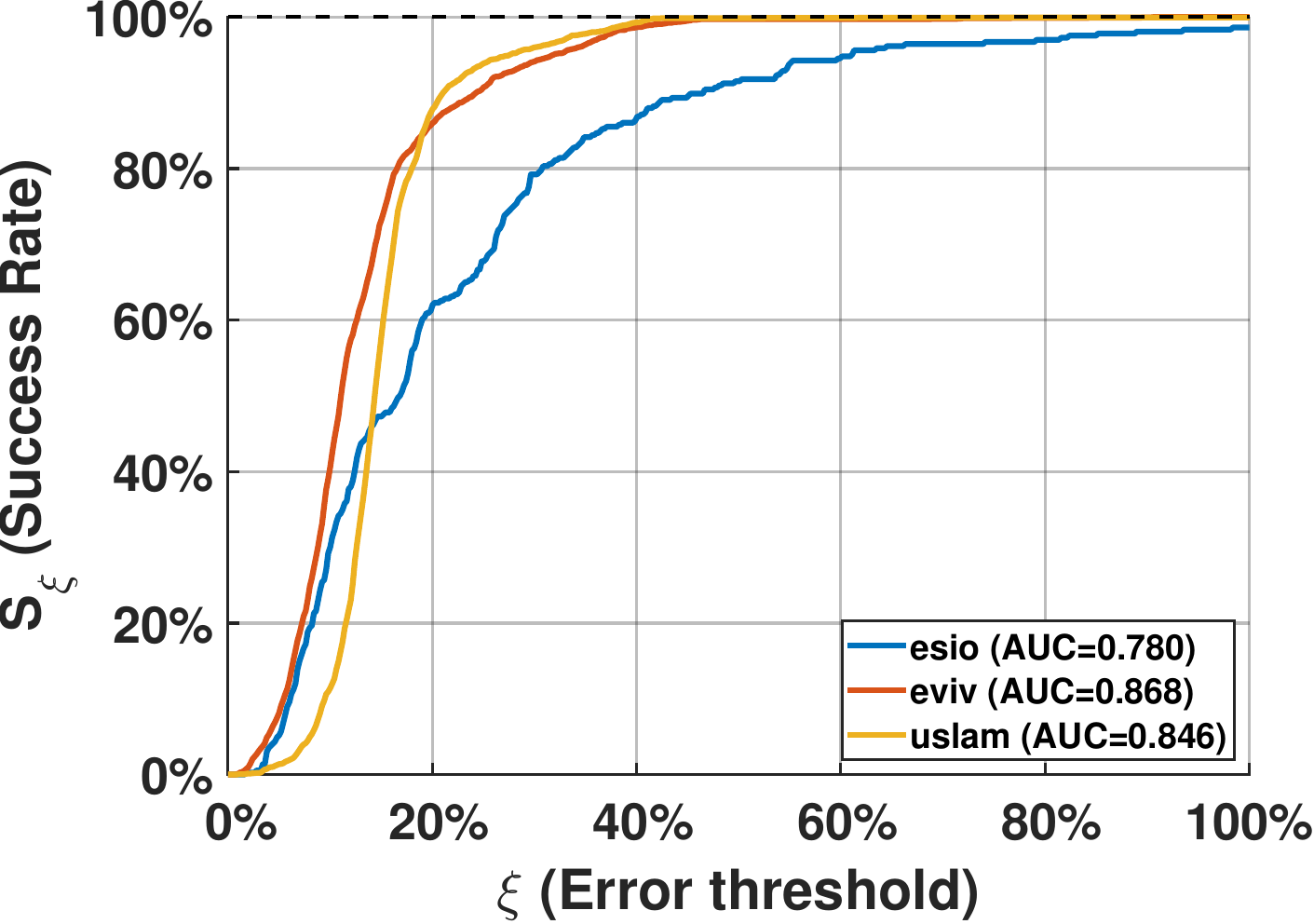}\label{subfig:unweighted_sr_curve}}~
  \subfigure[\small{Weighted $S({\xi})$ curves.}]{
  \includegraphics[width=0.20\textwidth]{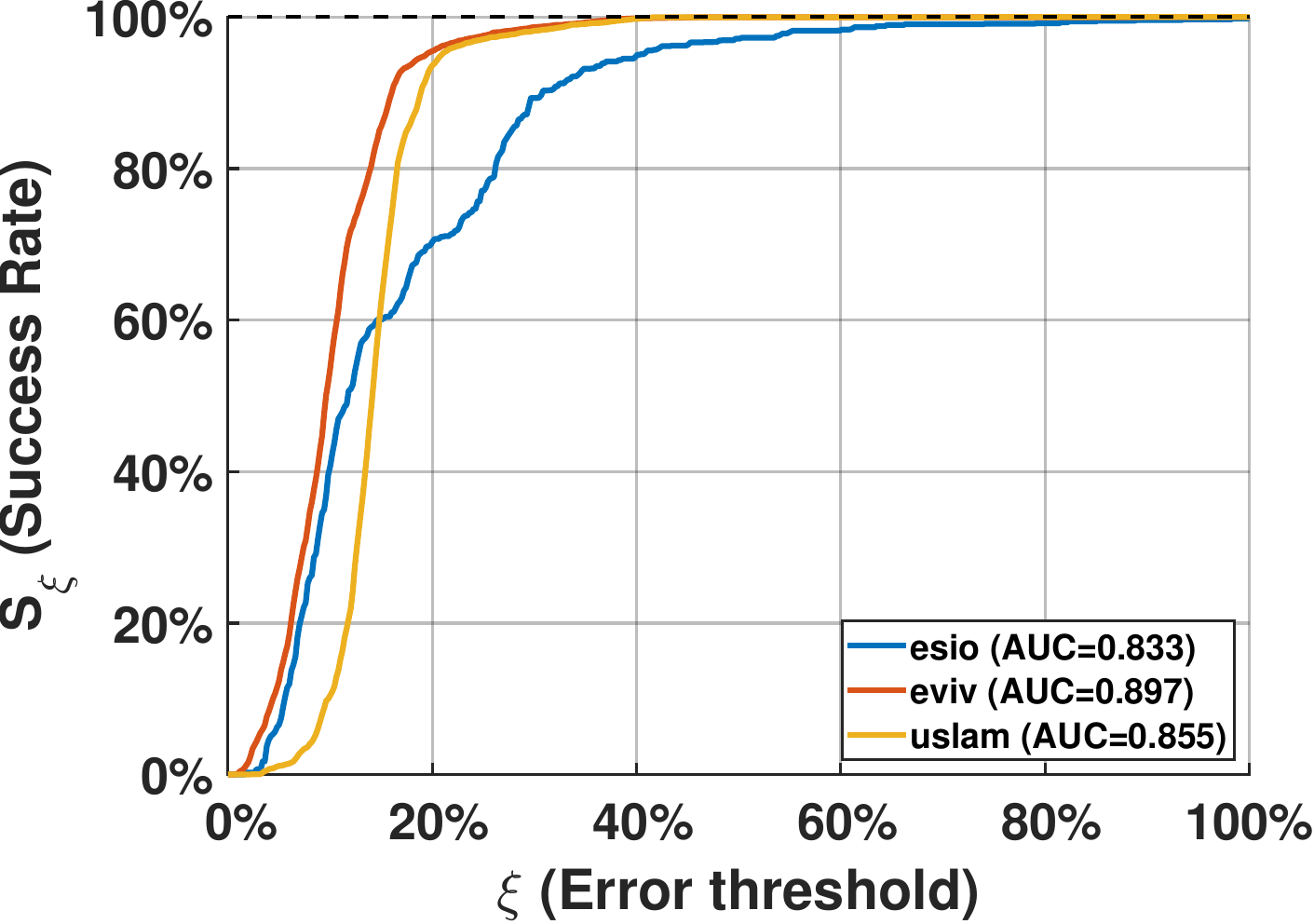}
  \label{subfig:weighted_sr_curve}}
 \caption{\textit{Comparison of different evaluation methods}.
 (a): RVE statistics of three methods (ESIO~\cite{Chen23ral}, EVIV~\cite{Lu24rss}, and USLAM~\cite{Rosinol18ral}), obtained using raw data from \cite{Lu24rss};
 (b): Boxplot results of the RVE statistics shown in (a);
 (c): Unweighted relative-velocity precision 
 curves (\ie, \eqref{eq:novel:metric} with $w_i=1/n$);
 and (d): 
 Relative-velocity precision curves using \eqref{eq:novel:metric}.
 \label{fig: evaluaiton stategies comparison}}
 \end{figure*}
 
Accurate velocity estimation is critical for closed-loop control systems, where rapid and precise responses are required. 
However, there is a lack of a standardized metric specifically designed for the assessment of linear velocity error.
While the translation component of RPE could serve as an approximate measure of velocity error, this strategy suffers from inherent discretization errors caused by finite sampling intervals. 
%
This issue could be resolved by computing either the Absolute Velocity Error (AVE) between the estimated and ground truth velocities or the corresponding Relative Velocity Error (RVE):
\begin{equation}
\begin{split}
\text{AVE}_i &= {\| \bfv_{\text{gt},i} - \bfv_{\text{est},i} \|_2 }, \\
\text{RVE}_i &= \text{AVE}_i \, / \, \| \bfv_{\text{gt},i} \|_2 .
\end{split}
\label{eq:evaluate metric}
\end{equation}
Then velocity errors can be probabilistically characterized by statistics (\eg,~mean, median, standard deviation) and visualized via box plots.
While this approach is simple, it has a key limitation: 
it treats all errors uniformly, failing to distinguish between cases with identical velocity errors that occur at different speeds. 
This situation is a critical drawback, as identical error values reflect fundamentally different performance levels in low-speed versus high-speed scenarios.

Inspired by precision-recall curves \cite{Pizzoli14icra}, 
we present a novel metric for quantitatively evaluating linear velocity estimation. 
Let the analog of the precision curve for velocity be the percentage of relative errors that fall below a certain threshold~$\xi$,
\begin{align}
\label{eq:novel:metric}
    S({\xi}) \doteq \sum_{i=1}^{n} w_i\; \mathbf{1}(\text{RVE}_i <\xi),
\end{align}
where 
$\mathbf{1}(x)$ denotes the indicator function 
and we define weights $w_i \doteq \|\mathbf{v}_{\text{gt},i} \| / {\sum_{j=1}^{n}\| \mathbf{v}_{\text{gt},j} \|}$.
In general, the proposed metric evaluates the success rate of velocity estimation, defined by the occurrence of the $\text{RVE}_i$ lower than a threshold $\xi$.

Note that RVE is used instead of AVE, because AVE ignores speed dependency, masking performance differences across operating ranges.
Also note that the weights $w_i$, defined by the magnitude of the linear velocity, highlight different performance levels in low-speed versus high-speed scenarios.

As a result, the curve \eqref{eq:novel:metric} can be drawn, and the \emph{Area Under the Curve} (AUC) is used as a unified statistic to rank the performance of different methods.
An illustrative comparison of the above mentioned evaluation strategies is provided in Fig.~\ref{fig: evaluaiton stategies comparison}.
Note that the weighted precision curves (Fig.~\ref{subfig:weighted_sr_curve}) shift leftward relative to their unweighted ($w_i=1/n$) counterparts (Fig.~\ref{subfig:unweighted_sr_curve}). 
This shows that successful estimates during high-speed motion contribute more significantly to the metric than those during low-speed motion, producing larger AUC values.

For scenarios with simultaneously high linear and angular velocities (e.g., orbital egomotion), 
we may use in \eqref{eq:novel:metric} weights that reflect the overall motion magnitude:
$$w_i \doteq \sqrt{ \left\| \mathbf{v}_{\text{gt},i} \right\|^2 + \left\| \boldsymbol{\omega}_{\text{gt},i} \right\|^2}
\left/ {\sum_{j=1}^{n} \sqrt{ \left\| \mathbf{v}_{\text{gt},j} \right\|^2 + \left\| \boldsymbol{\omega}_{\text{gt},j} \right\|^2}}\right..$$

\subsection{Comparison of State-of-the-Art Methods}

\begin{table*}[t]
\centering
\setlength{\tabcolsep}{4pt}
\caption{Evaluation results of the SOTA event-based SLAM methods on the first batch of sequences [ATE:(m), AUC:(-)]}
\label{tab:Evaluation}
\resizebox{\textwidth}{!}{
\renewcommand{\arraystretch}{1.2}
\begin{tabular}{l*{11}{c}}
\toprule
Modality & \multicolumn{3}{c}{\textbf{Events and Frames (E + F)}} & \multicolumn{5}{c}{\textbf{E-only}} & \multicolumn{3}{c}{\textbf{F-only}}\\
\cmidrule(lr){2-4}
\cmidrule(lr){5-9}
\cmidrule(lr){10-12}
\multirow{2}{*}{Method} & FETAP~\cite{Liu25iros} & \makecell{SuperEvent~\cite{Burkhardt25iccv}} & REFIO~\cite{refio} & \makecell{$\text{SuperEvent}^{*}$~\cite{Burkhardt25iccv}} & SDEVO~\cite{Zhong25ral} & DEIO~\cite{Guan25iccvw} & DEVO~\cite{Klenk24threedv} & EVIV~\cite{Lu25tro} & VINS~\cite{Qin18tro} & OKVIS2~\cite{Leutenegger22okvis2} & DPVO~\cite{Teed23dpvo} \\
{~} & Mono EVIO & Stereo EVIO & Stereo EVIO & Stereo EIO & Stereo EO & Mono EIO & Mono EO & Stereo EIO & Stereo VIO & Stereo VIO & Mono VO \\
\cmidrule(lr){2-4}
\cmidrule(lr){5-9}
\cmidrule(lr){10-12}
& ATE / AUC & ATE / AUC & ATE / AUC & ATE / AUC & ATE / AUC & ATE / AUC & ATE / AUC & ATE / AUC & ATE / AUC & ATE / AUC & ATE / AUC \\
\midrule
\emph{drone\_circle\_norm} & 0.292 / \textbf{0.890} & 0.109 / \textbf{0.890} & 0.273 / 0.875 & 0.189 / 0.879 & 0.373 / 0.860 & \textbf{0.067} / 0.885 & 1.158 / 0.841 & 2.068 / 0.872 & 0.435 / 0.874 & 0.144 / 0.826 & 0.190 / 0.826 \\
\emph{drone\_circle\_fast} & 0.113 / 0.899 & 0.095 / \textbf{0.902} & 0.298 / 0.888 & 0.282 / 0.883 & 0.388 / 0.871 & \textbf{0.069} / 0.894 & 1.230 / 0.847 & 1.659 / 0.872 & 0.186 / 0.898 & 0.151 / 0.829 & 0.267 / 0.830 \\
\emph{drone\_line\_norm} & 0.116 / 0.930 & 0.035 / 0.926 & 0.310 / 0.895 & 0.080 / 0.904 & 0.418 / 0.905 & 0.043 / \textbf{0.933} & 0.137 / 0.909 & 0.885 / 0.879 & 0.179 / 0.924 & 0.036 / 0.915 & \textbf{0.031} / 0.921 \\
\emph{drone\_line\_fast} & 0.080 / \textbf{0.911} & 0.060 / \textbf{0.911} & 0.168 / 0.870 & \textbf{0.055} / 0.901 & 0.344 / 0.875 & 0.066 / \textbf{0.911} & 0.287 / 0.897 & 2.283 / 0.808 & 0.114 / 0.907 & 0.059 / 0.873 & 0.062 / 0.877 \\
\emph{slider\_norm} & \textbf{0.019} / \textbf{0.921} & 0.041 / 0.881 & 0.083 / 0.899 & 0.045 / 0.869 & 0.038 / 0.914 & 0.077 / 0.856 & \textbf{0.019} / 0.919 & 0.227 / 0.861 & 0.080 / 0.845 & 0.145 / 0.758 & 0.068 / 0.865 \\
\emph{arm\_norm} & 0.048 / 0.831 & \textbf{0.033} / \textbf{0.857} & 0.164 / 0.784 & \textbf{0.033} / \textbf{0.857} & 0.069 / 0.855 & 0.078 / 0.655 & 0.054 / 0.816 & 1.966 / 0.805 & 0.170 / 0.694 & 0.708 / 0.653 & 0.059 / 0.730 \\
\emph{car\_circle\_norm} & 0.149 / 0.904 & 0.076 / 0.904 & 0.131 / 0.873 & 0.179 / 0.844 & 0.263 / 0.854 & \textbf{0.064} / \textbf{0.913} & 0.343 / 0.854 & 1.505 / 0.784 & 0.150 / 0.887 & 0.229 / 0.814 & 0.215 / 0.815 \\
\midrule
Average & 0.117 / \textbf{0.898} & \textbf{0.064} / 0.896 & 0.192 / 0.869 & 0.123 / 0.877 & 0.271 / 0.876 & 0.066 / 0.864 & 0.461 / 0.869 & 1.513 / 0.840 & 0.188 / 0.861 & 0.210 / 0.810 & 0.127 / 0.838 \\
\bottomrule
\end{tabular}}
\end{table*}
\begin{table}[t] 
\centering
\setlength{\tabcolsep}{4pt} 
\caption{Results of SOTA event-based SLAM methods on the second batch of sequences [ATE:(m), AUC:(-)]. 
}
\label{tab:Evaluation2}
\resizebox{\columnwidth}{!}{
\begin{tabular}{l*{4}{c}} 
\toprule
Method & \makecell{ESVIO~\cite{Chen23ral}} & \makecell{SDEVO~\cite{Zhong25ral}} & \makecell{VINS~\cite{Qin18tro}} & \makecell{DPVO~\cite{Teed23dpvo}} \\
& Stereo EVIO & Stereo EO & Stereo VIO & Mono VO \\
\cmidrule(l{1mm}r{1mm}){2-2}
\cmidrule(l{1mm}r{1mm}){3-3}
\cmidrule(l{1mm}r{1mm}){4-4}
\cmidrule(l{1mm}r{1mm}){5-5}
Sequence & ATE / AUC & ATE / AUC & ATE / AUC & ATE / AUC \\
\midrule
\emph{drone\_8\_norm} & 0.403 / 0.934 & 0.881 / 0.880 & 1.031 / 0.921 & \textbf{0.333} / \textbf{0.943} \\
\emph{drone\_8\_hdr} & 0.876 / 0.922 & 1.935 / 0.818 & 1.732 / 0.767 & \textbf{0.352} / \textbf{0.928} \\
\emph{drone\_s\_norm} & \textbf{0.639} / \textbf{0.899}  & 1.502 / 0.772 & 1.029 / 0.759 & 1.743 / 0.674 \\
\emph{drone\_s\_hdr} & \textbf{0.445} / \textbf{0.822} & 2.044 / 0.691 & 2.425 / 0.744 & 2.300 / 0.518 \\
\emph{drone\_rot\_norm} & \textbf{0.386} / \textbf{0.752}  & 0.421 / 0.577 & 0.664 / 0.505 & 0.673 / 0.402 \\
\emph{drone\_rot\_fast} & 0.742 / \textbf{0.563} & 0.549 / 0.374 & 2.365 / 0.283 & \textbf{0.431} / 0.529 \\
\emph{slider\_fast} & 0.381 / 0.601  & \textbf{0.115} / \textbf{0.654} & 0.730 / 0.382 & 0.147 / 0.623 \\
\emph{slider\_hdr} & 0.278 / 0.544  & 0.197 / 0.728 & 0.377 / 0.572 & \textbf{0.124} / \textbf{0.777} \\
\emph{arm\_dim} & 0.137 / 0.637 & 0.261 / 0.602 & 0.341 / 0.588 & \textbf{0.066} / \textbf{0.678}  \\
\emph{arm\_hdr} & 0.219 / 0.594 & 0.512 / 0.531 & 0.600 / 0.488 & \textbf{0.087} / \textbf{0.660} \\
\emph{car\_circle\_hdr} & 1.127 / 0.541  & 1.713 / 0.669 & 0.842 / 0.513 & \textbf{0.380} / \textbf{0.704} \\
\midrule
Average & \textbf{0.512} / \textbf{0.710} & 0.921 / 0.663 & 1.103 / 0.593 & 0.603 / 0.676 \\
\bottomrule
\end{tabular}}
\end{table}

\begin{figure*}[t]
	\centering

    \subfigure[\emph{drone\_circle\_fast}]{
        \centering
        \includegraphics[width=0.465\textwidth]{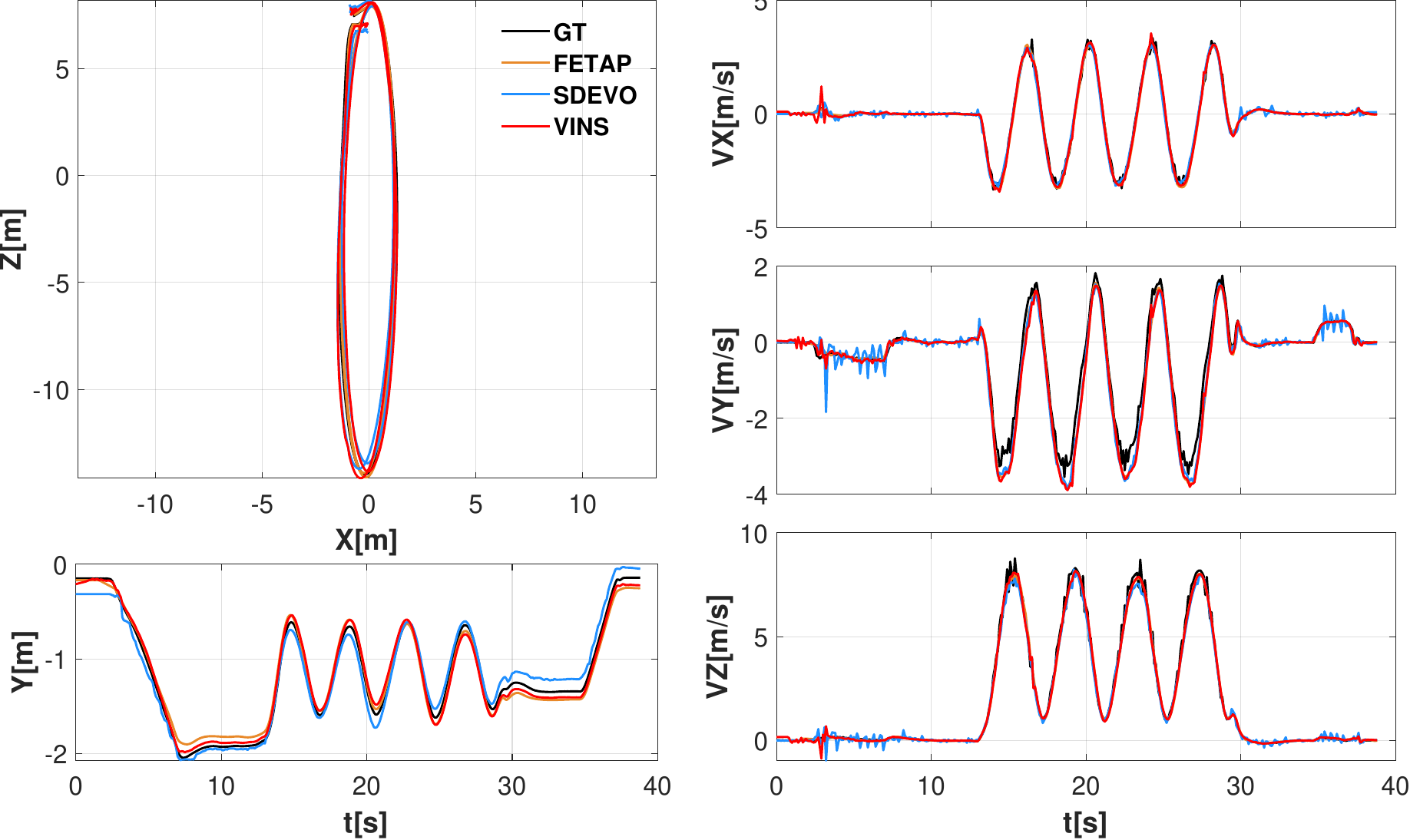}
        \label{subfig:drone_circle_fast}
    }\qquad
	\subfigure[\emph{arm\_norm}]{
        \centering
        \includegraphics[width=0.465\textwidth]{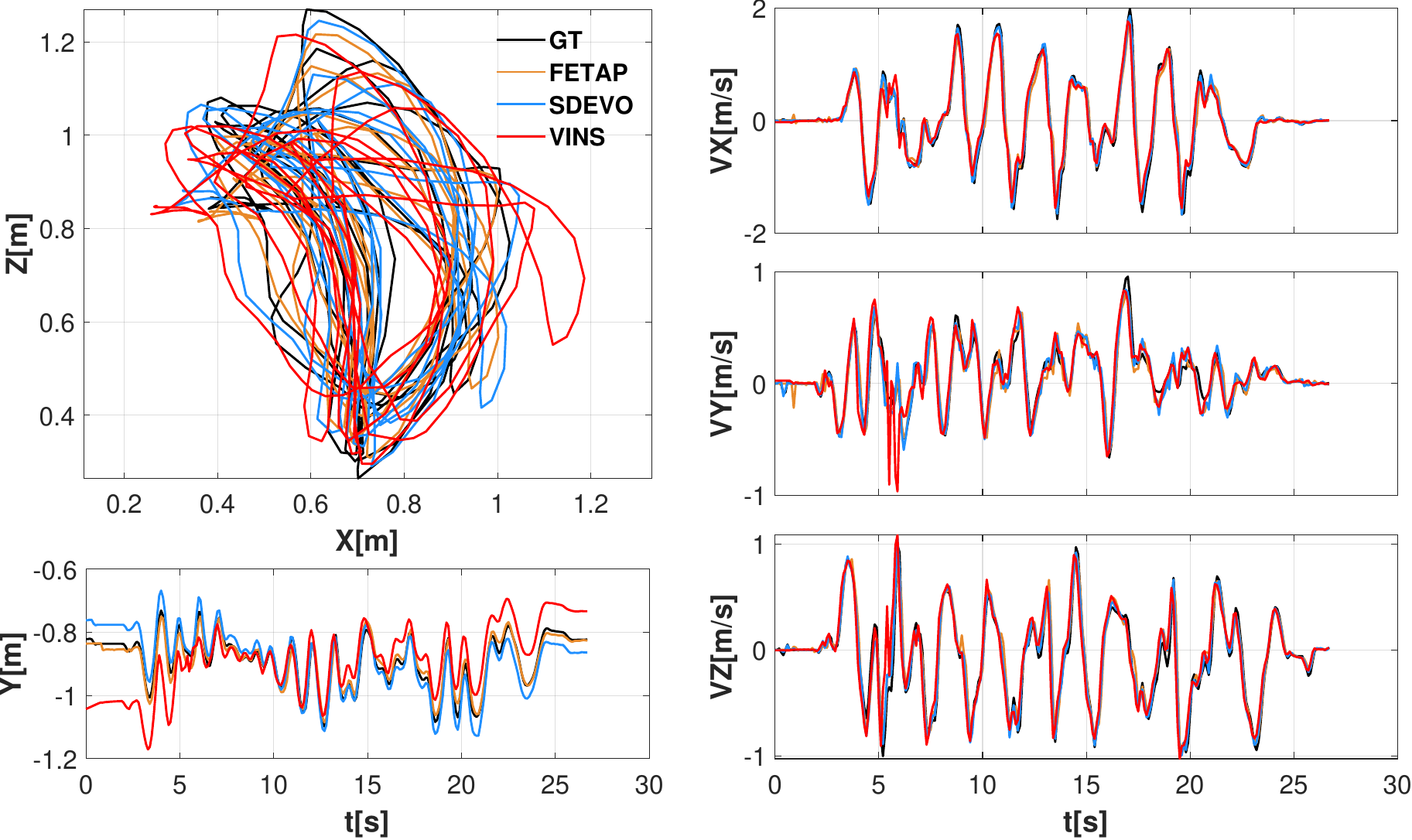}
        \label{subfig:arm_normal}
    } \\
    \subfigure[\emph{drone\_8\_hdr}]{
        \centering
        \includegraphics[width=0.465\textwidth]{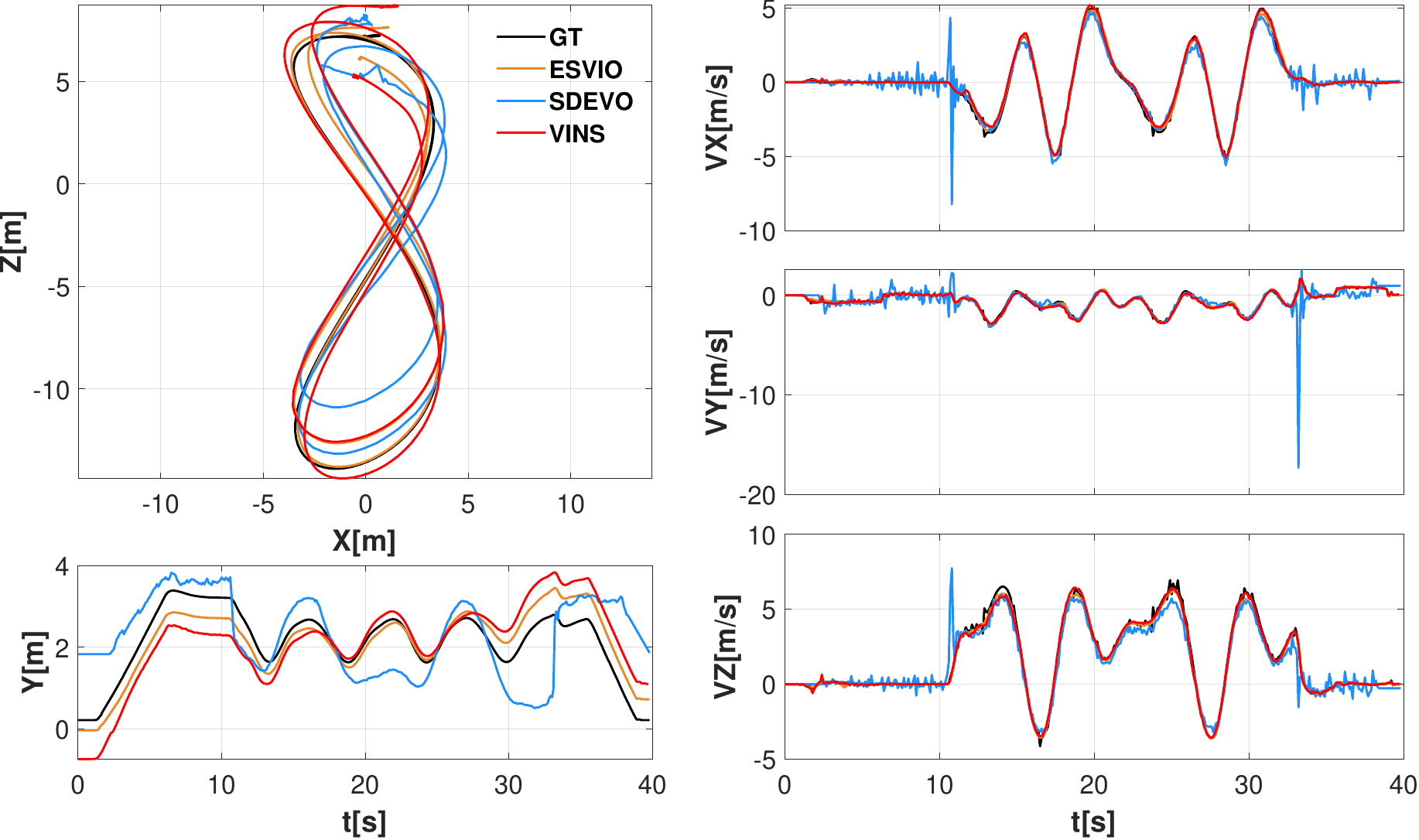}
        \label{subfig:drone_8_hdr}
    }\qquad
	\subfigure[\emph{drone\_s\_hdr}]{
        \centering
        \includegraphics[width=0.465\textwidth]{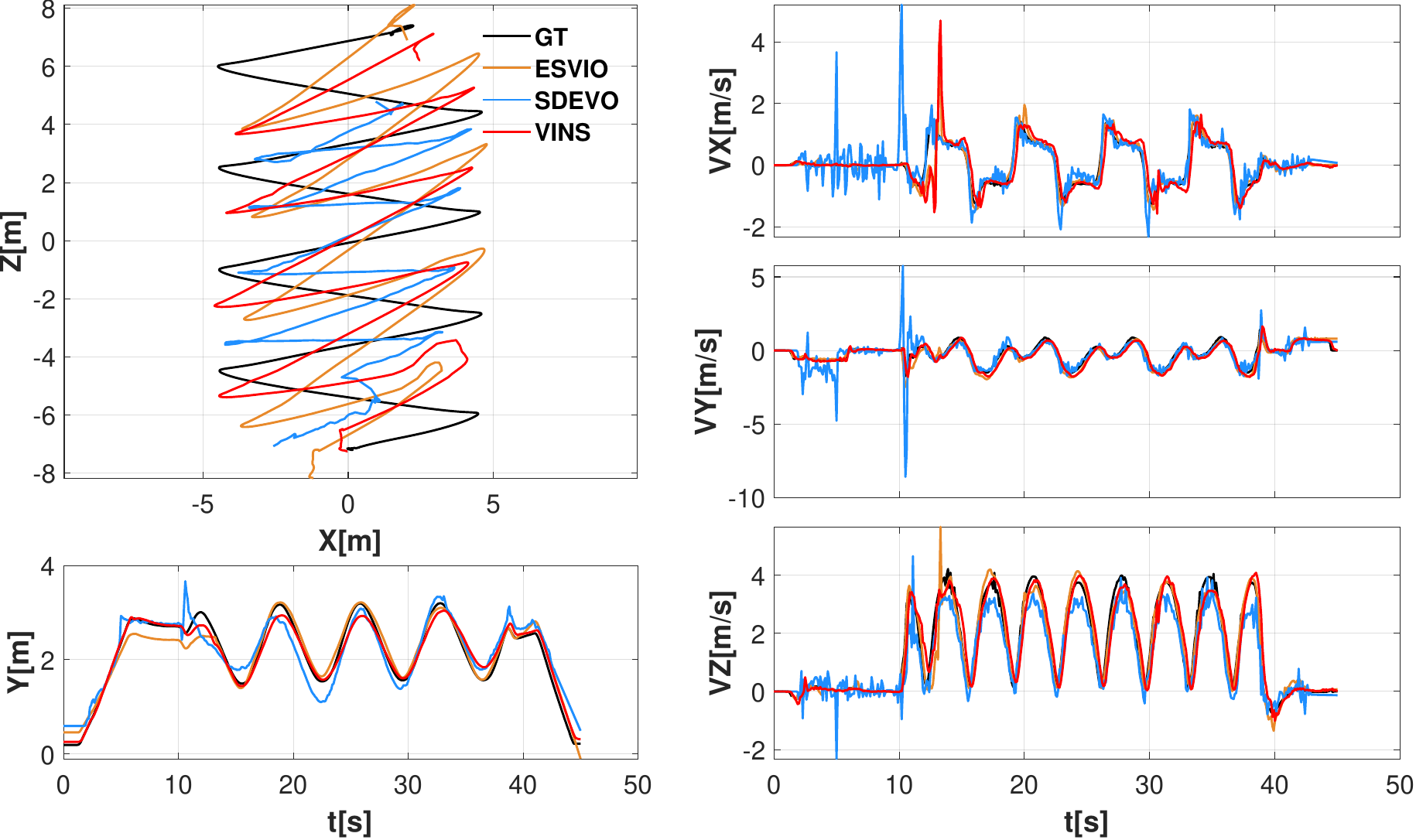}
        \label{subfig:drone_s_hdr}
    } \\
    \subfigure[\emph{slider\_hdr}]{
        \centering
        \includegraphics[width=0.465\textwidth]{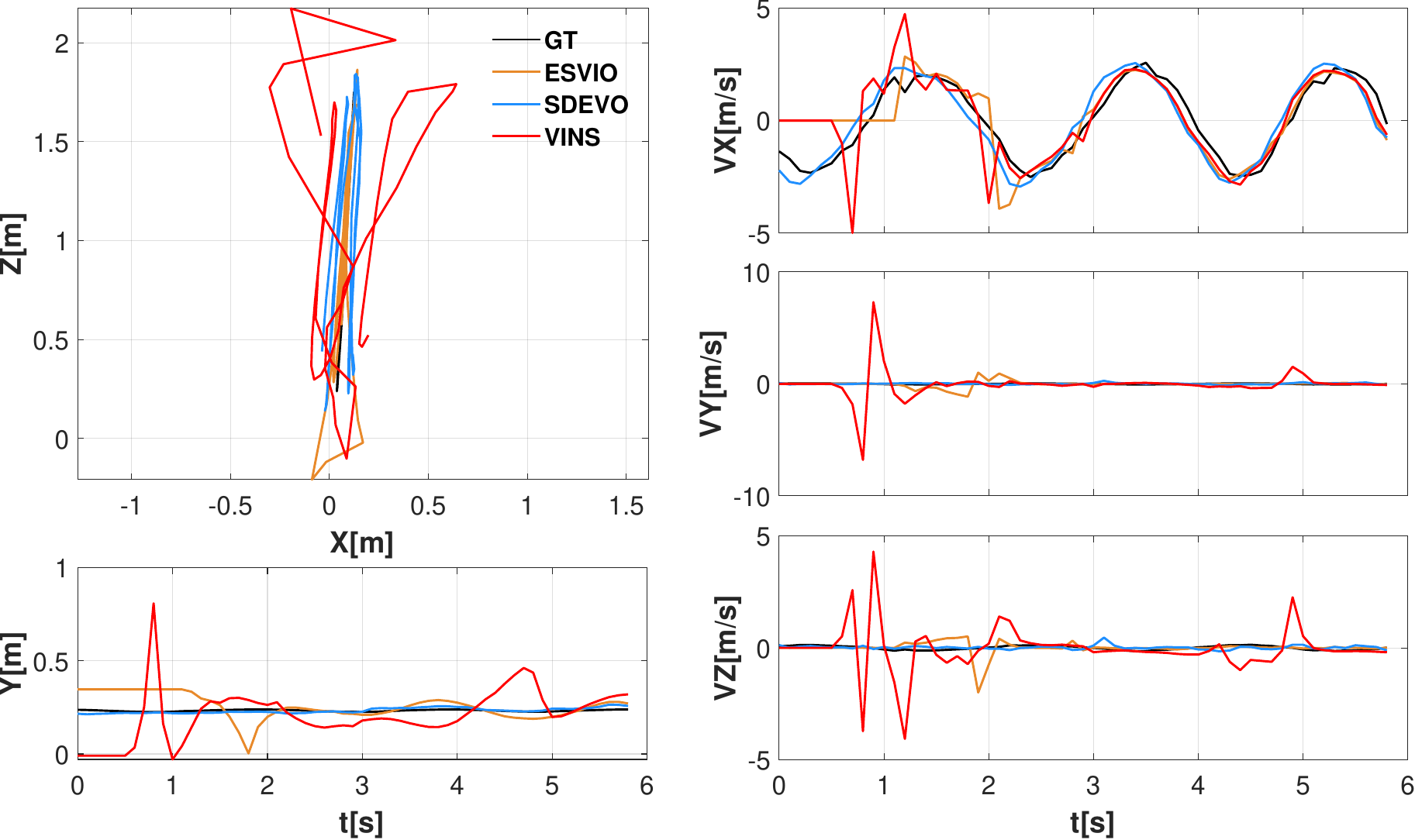}
        \label{subfig:slider_s_hdr}
    }\qquad
	\subfigure[\emph{car\_circle\_hdr}]{
        \centering
        \includegraphics[width=0.465\textwidth]{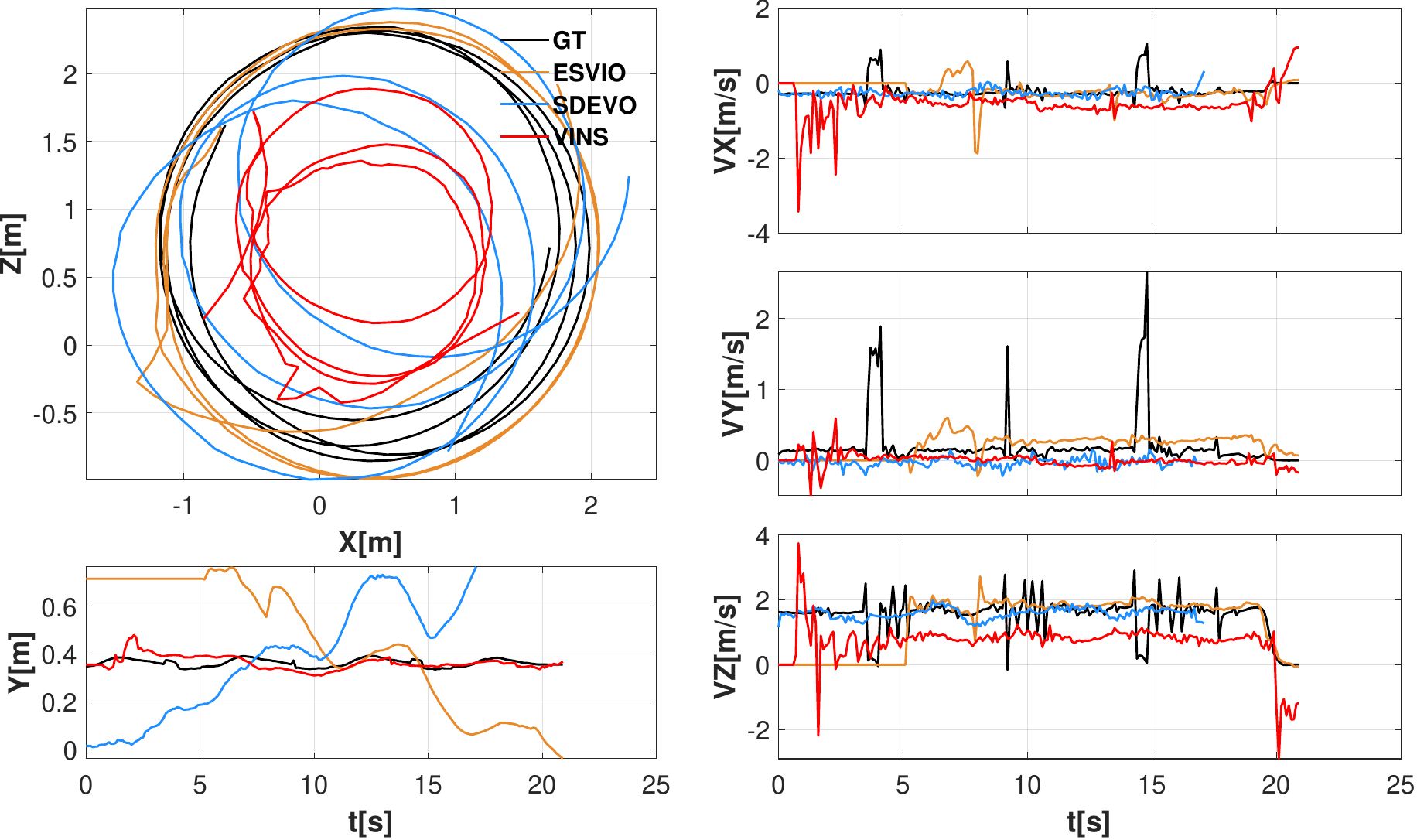}
        \label{subfig:car_circle_hdr}
    } 
	\caption{\textit{Visualization of part evaluated methods on selected sequences.} 
    In each sequence, the left plots show bird’s-eye view of trajectories and details along the $Y$-axis;
    and the right plots report velocity ($X,Y,Z$ axes).} 
    \label{fig:traj_vel}
\end{figure*}
We evaluate multiple state-of-the-art event-based approaches, including both model-based and deep learning methods, 
on the proposed EvSLAM datasets. 
Since some methods have not been formally named or published, we refer to them by the names of their key modules:
\begin{itemize}
    \item FETAP is a monocular event-image method, which first detects uniformly distributed keypoints on image frames using SuperPoint~\cite{Detone18cvprw}, then employs the FETAP algorithm~\cite{Liu25iros} to fuse the event stream with image frames for high-frequency feature tracking, generating smooth and continuous feature trajectories.
    These trajectories are then fed into the backend of the VINS-mono~\cite{Qin18tro} framework for tightly-coupled optimization with inertial measurement to estimate pose.

\item SuperEvent~\cite{Burkhardt25iccv} adopts a stereo setup, which uses a transformer backbone to detect keypoints and generate descriptors from multi-channel event representations. 
This module is integrated into OKVIS2~\cite{Leutenegger22okvis2}, a state-of-the-art frame-based VIO framework, where it operates in parallel with or directly replaces the frame-based BRISK~\cite{Leutenegger11iccv} keypoint detection module (distinguished by *).

\item Similarly, ESVIO~\cite{Chen23ral} builds upon the classical feature-based pipeline of VINS-Fusion~\cite{Qin18tro} by detecting features from motion-compensated event representations using Arc*~\cite{Alzugaray18ral} corner detectors, and tracking them with pyramidal KLT optical flow~\cite{Lucas81ijcai} to construct event-based feature constraints for joint optimization.

\item REFIO \cite{refio} follows the same overall strategy as ESVIO, while further exploiting geometric constraints constructed from event features for scale refinement and adopting an adaptive weighting strategy during backend optimization to balance the contributions of the two visual measurements, thereby effectively leveraging both modalities.

\item SDEVO~\cite{Zhong25ral} and DEIO~\cite{Guan25iccvw} are both derived from DEVO~\cite{Klenk24threedv}, which uses a recurrent network to estimate sparse patch-based optical flow and integrate it into a differentiable bundle adjustment layer for pose estimation. 
SDEVO recovers metric scale by incorporating stereo observations, while DEIO introduces inertial constraints for scale estimation and additionally employs a proximity-based loop closure to reduce accumulated drift.

\item In contrast to the above fixed-frame-rate approaches, EVIV~\cite{Lu25tro} asynchronously estimates normal flow from the event stream, then fuses normal flow constraints with inertial measurements in a continuous-time optimization framework to estimate linear velocity.

\textcolor{black}{\item To more clearly demonstrate the advantages of event cameras over frame-based cameras under high-speed maneuvers and challenging illumination conditions, we evaluate three frame-based methods: VINS-Fusion~\cite{Qin18tro} in its stereo visual-inertial configuration, the optimization-based stereo visual-inertial SLAM system OKVIS2~\cite{Leutenegger22okvis2}, and the learning-based monocular visual odometry method DPVO~\cite{Teed23dpvo}. 
We additionally evaluate DEVO~\cite{Klenk24threedv}, an event-based monocular visual odometry method built upon DPVO, to enable a comparison using the same monocular configuration.
Since both DPVO and DEVO exhibit scale ambiguity, we apply scale alignment to their estimated trajectories prior to evaluation.}
\end{itemize}

The evaluation results of multiple methods on the first and second batches of sequences are presented in Tabs.~\ref{tab:Evaluation} and~\ref{tab:Evaluation2}, respectively. 
\textcolor{black}{The evaluated methods differ between the two batches because open-source implementations are unavailable for some methods.}
Visualizations are provided in Fig.~\ref{fig:traj_vel}. 
Three conclusions can be drawn from these results. 
First, the continuous constraints provided by the IMU can improve the smoothness of linear velocity estimation.
This is verified by the comparison between DEIO and SDEVO, as well as between VINS-Fusion and DPVO: systems equipped with an IMU exhibit significantly higher AUC values on drone sequences with high acceleration, even if their ATE is potentially larger.

Second, event cameras outperform frame-based cameras in high-speed maneuvering scenarios, and this advantage becomes increasingly pronounced as the optical flow magnitude increases. 
It is worth noting that in the first batch of sequences, benefiting from the normal lighting conditions and the auto-exposure adjustment capability of the D435 camera, significant motion blur in grayscale images only occurs during occasional moments of high angular velocity, thus enabling all frame-based camera algorithms to achieve satisfactory performance.
\textcolor{black}{Compared with DEVO and its stereo extension SDEVO, DPVO achieves lower ATE on most sequences, benefiting from the rich texture information in grayscale images that remain clear during much of each sequence. However, both event-based methods achieve higher AUC than DPVO on most sequences, with the AUC advantage particularly pronounced in \emph{arm\_norm}, which belongs to the highest optical-flow tier in the first batch.}
The results of SuperEvent also corroborate this observation: in the \emph{arm\_norm} sequence, whether a frame-based camera is used has no impact on the final results.
However, the situation changes when low-light and HDR scenarios are introduced in the second batch of sequences.
Although event cameras can capture textures in extremely bright regions under such conditions (e.g., \emph{slider\_hdr} in Fig.~\ref{fig: sequences_overview}), event data quality deteriorates significantly under high-speed maneuvers.
This is clearly illustrated by the comparison between \emph{drone\_8\_norm} and \emph{drone\_8\_hdr} in Fig.~\ref{fig: sequences_overview}: the latter presents only a few clear event edges and is dominated by substantial noise, a phenomenon commonly observed in HDR sequences.
In contrast, grayscale images remain reasonably discriminative despite being more blurred.
Consequently, the event-only method SDEVO achieves noticeably lower accuracy than DPVO on the second dataset, particularly in low-light and HDR sequences.
We believe that this observation suggests potential limitations of event cameras in scenarios characterized by the simultaneous presence of HDR and high-speed maneuvers.

Building upon the second conclusion, we can naturally draw the following one: frame-based cameras and event cameras can complement each other to achieve more accurate pose and velocity estimation. 
Frame-based cameras provide abundant texture information to ensure high-precision pose estimation in medium-to-low speed scenarios, while the event cameras leverage their high temporal resolution and high dynamic range to overcome interference in high-speed maneuvering scenarios.
The two sets of experimental results for SuperEvent provide a clear example: simply incorporating image feature constraints together with event feature constraints for joint optimization leads to improved performance on most sequences. 
Similarly, compared to VINS-Fusion, REFIO achieves significant performance gains on the challenging \emph{arm\_norm} sequence by introducing event-based feature constraints. 
An analogous but even more compelling comparison can be found in the second batch of data between ESVIO and VINS-Fusion, where the former consistently outperforms the latter in terms of both ATE and AUC across all sequences, even in HDR scenarios where the quality of event data is significantly degraded.

\subsection{Runtime Analysis}
To verify our conclusion in Sec.~\ref{subsec: resolution} that VGA resolution is currently the most suitable one for event-based SLAM, and to more comprehensively demonstrate the performance of state-of-the-art event-based methods, we report the runtime of several representative approaches in Tab.~\ref{tab:runtime}, where the evaluations are conducted on the self-recorded \emph{drone\_circle\_norm} sequence at VGA resolution and the \emph{mocap\_6dof} sequence from the TUM-VIE~\cite{Klenk21iros} dataset at HD resolution.
As some methods are not open-source or are only partially available, we can only measure the runtime of their core modules on different platforms.
Specifically, FETAP is evaluated on a desktop computer equipped with an Intel Core i7-14700K CPU and a RTX 3090 GPU, while the other methods are tested on a desktop configuration featuring an Intel Core i7-14700K CPU and a RTX 4080 GPU.
For the partially available method SuperEvent, only the runtime of event representation, feature detection and matching is reported.
Since these methods adopt diverse system architectures, we perform a coarse decomposition of each system into a pre-processing module for event representation (which adapt the input events to the algorithms), a front-end responsible for constructing event-based feature constraints and an optimization back-end, without accounting for implementation details, such as whether the front-end and back-end run within the same thread (e.g., SDEVO) or whether multiple threads are employed in the front-end (e.g., ESVIO and EVIV).
It is worth noting that FETAP employs a network to generate event representations at up to 200 Hz, which explains why its pre-processing runtime is so short and remains almost unchanged across different resolutions.

The values in Tab.~\ref{tab:runtime} show that ($i$) runtime is generally dominated by pre-processing and front-end, 
($ii$) and in these two components the computational cost of HD is consistently larger than VGA, as expected by the larger resolution and number of events (see Tab.~\ref{tab:event_statistics}).

Although Tab.~\ref{tab:runtime} suggests that some current data-driven deep learning methods demonstrate a certain level of real-time performance on VGA-resolution event data, even outperforming some model-based methods, it is important to note that these methods heavily rely on high-performance GPUs. 
When deployed on power-constrained embedded platforms, their computational latency increases significantly \cite{Zhong25ral}. 
In addition, although the front-end runtime of learning-based methods is generally not sensitive to increasing resolution, the cost of constructing the high-dimensional and complex event representations used to extract implicit feature information is often linearly correlated with the event count. 
More importantly, the potentially substantial overhead required for transmitting such event representations is not included in the present analysis, which is also closely related to the resolution.
%
Consequently, a considerable gap remains between these methods and the real-time performance required for practical applications, while higher-resolution event data remain far beyond the capabilities of current event-based SLAM methods.
\begin{table}[t]
\begin{center}
\renewcommand\arraystretch{1.2}
\caption{Runtime of several methods. [Time: ms]}
\label{tab:runtime}
\begin{adjustbox}{max width=\linewidth}
\setlength{\tabcolsep}{5pt}{
\begin{tabular}{lrrrrrr}
\toprule
\textbf{Method} & \multicolumn{2}{c}{\textbf{Pre-processing}} & \multicolumn{2}{c}{\textbf{Front-end}} & \multicolumn{2}{c}{\textbf{Back-end}} \\
\cmidrule(lr){2-3}
\cmidrule(lr){4-5}
\cmidrule(lr){6-7}
& \multicolumn{1}{c}{VGA} & \multicolumn{1}{c}{HD} 
& \multicolumn{1}{c}{VGA} & \multicolumn{1}{c}{HD} 
& \multicolumn{1}{c}{VGA} & \multicolumn{1}{c}{HD} \\
\midrule
FETAP \cite{Liu25iros} & 3.02 & 4.91 &  151.05 & 161.25 & 25.20 & 23.27 \\
$\text{SuperEvent}^{*}$ \cite{Burkhardt25iccv} & 147.75 & 723.47 & 13.76 & 40.98 & - & -\\
SDEVO \cite{Zhong25ral} & 22.14 & 168.90 & 17.14 & 26.33 & 1.23 & 1.25\\
ESVIO \cite{Chen23ral} & 33.99 & 61.09 & 5.32 & 17.80 & 32.55 & 29.56\\
EVIV \cite{Lu25tro} & 91.80 & 703.19 & 107.14 & 126.22 & 6.29 & 6.19 \\
\bottomrule
\end{tabular}
}
\end{adjustbox}

\par\smallskip
\begin{minipage}{\linewidth}
{\emph{* ``-'' represents the lack of experimental results of the algorithm }}
\end{minipage}

\end{center}
\end{table}

\section{Conclusion}
\label{sec: conclusion}
In this paper, we present EvSLAM, a comprehensive multi-platform high-speed maneuvering benchmark designed to quantitatively assess the extent to which the potential of event cameras has been fully unlocked. 
We highlight and analyze the shortcomings in current publicly available event-based SLAM datasets, including spatio-temporal inconsistencies across stereo observations, the lack of comprehensive high-speed maneuvering scenarios, and the computational burden imposed by excessively high resolutions. 
Building on these insights, we record multiple challenging high-speed maneuvering sequences encompassing both standard and complex lighting conditions, using a multi-camera configuration with VGA resolution, which includes a stereo event camera, stereo grayscale cameras, and a color camera, on four common mobile robotic platforms: a drone, a sliding rail, a robotic arm, and a Mecanum-wheel vehicle.
All sequences are annotated with high-precision ground-truth data. 
By evaluating these sequences with state-of-the-art event-based VO/SLAM methods and two classic frame-based approaches, and combining the results with a review of recent advancements in event-based SLAM, we explore the characteristics and fundamental limitations of different categories of event-based SLAM methods and discuss future directions of the field.
We believe that this benchmark can serve as a foundation for exploring the potential of event cameras in extreme scenarios, such as high-speed maneuvers.



\end{document}